\documentclass[runningheads]{llncs}
\usepackage[table,xcdraw]{xcolor}
\usepackage{graphicx}

\usepackage{tikz}
\usepackage{comment}
\usepackage{amsmath,amssymb} 
\usepackage{color}
\usepackage{caption}
\usepackage{subcaption}

\usepackage[pagebackref,breaklinks,colorlinks,citecolor=blue,linkcolor=blue]{hyperref}
\usepackage{cite}
\newcommand{\etal}{\textit{et al.}}

\usepackage[accsupp]{axessibility}  

\usepackage{graphicx}
\usepackage{amsmath}
\usepackage{amssymb}
\usepackage{booktabs}
\usepackage{stackengine}
\usepackage{multirow}
\usepackage[capitalize]{cleveref}
\crefname{section}{Sec.}{Secs.}
\Crefname{section}{Section}{Sections}
\Crefname{table}{Table}{Tables}
\crefname{table}{Tab.}{Tabs.}

\begin{document}
\pagestyle{headings}
\mainmatter
\def\ECCVSubNumber{5129}  

\title{3D-PL: Domain Adaptive Depth Estimation with 3D-aware Pseudo-Labeling} 

\titlerunning{3D-PL: Domain Adaptive Depth Estimation with Pseudo-Labeling}
%
\author{Yu-Ting Yen$^{1,2}$, Chia-Ni Lu$^1$, 
Wei-Chen Chiu$^1$, Yi-Hsuan Tsai$^2$}
\authorrunning{Yen et al.}
%
\institute{$^1$National Chiao Tung University, Taiwan 
$^2$Phiar Technologies}
\maketitle

\begin{abstract}
For monocular depth estimation, acquiring ground truths for real data is not easy, and thus domain adaptation methods are commonly adopted using the supervised synthetic data. However, this may still incur a large domain gap due to the lack of supervision from the real data. In this paper, we develop a domain adaptation framework via generating reliable pseudo ground truths of depth from real data to provide direct supervisions. Specifically, we propose two mechanisms for pseudo-labeling: 1) 2D-based pseudo-labels via measuring the consistency of depth predictions when images are with the same content but different styles; 2) 3D-aware pseudo-labels via a point cloud completion network that learns to complete the depth values in the 3D space, thus providing more structural information in a scene to refine and generate more reliable pseudo-labels. In experiments, we show that our pseudo-labeling methods improve depth estimation in various settings, including the usage of stereo pairs during training. Furthermore, the proposed method performs favorably against several state-of-the-art unsupervised domain adaptation approaches in real-world datasets. Our code and models are available at \url{https://github.com/ccc870206/3D-PL}.

\keywords{domain adaptation, monocular depth estimation, pseudo-labeling}
\end{abstract}

\section{Introduction}
\label{sec:intro}
Monocular depth estimation is an ill-posed problem that aims to estimate depth from a single image. Numerous supervised deep learning methods \cite{eigen2014depth, liu2015learning, cao2017estimating, fu2018deep, yin2019enforcing, jiao2018look} have made great progress in recent years. However, they need a large amount of data with ground truth depth, while acquiring such depth labels is highly expensive and time-consuming because it requires depth sensors such as LiDAR~\cite{geiger2012we} or Kinect~\cite{zhang2012microsoft}. Therefore, several unsupervised methods~\cite{garg2016unsupervised, godard2017unsupervised, godard2019digging, mahjourian2018unsupervised, zhan2018unsupervised, tosi2019learning} have been proposed, where these approaches estimate disparity from videos or binocular stereo images without any ground truth depth. Unfortunately, since there is no strong supervision provided, unsupervised methods may not do well under situations such as occlusion or blurring in object motion. To solve this problem, recent works use synthetic datasets since the synthetic image-depth pairs are easier to obtain and have more accurate dense depth information than real-world depth maps. However, there still exists domain shift between synthetic and real datasets, and thus many works use domain adaptation~\cite{kundu2018adadepth, lopez2020desc, zheng2018t2net, zhao2019geometry, chen2019crdoco} to overcome this issue.

\begin{figure}[t]
     \centering
     \resizebox{0.55\linewidth}{!}{
     \begin{subfigure}{0.5\linewidth}
         \includegraphics[width=\linewidth]{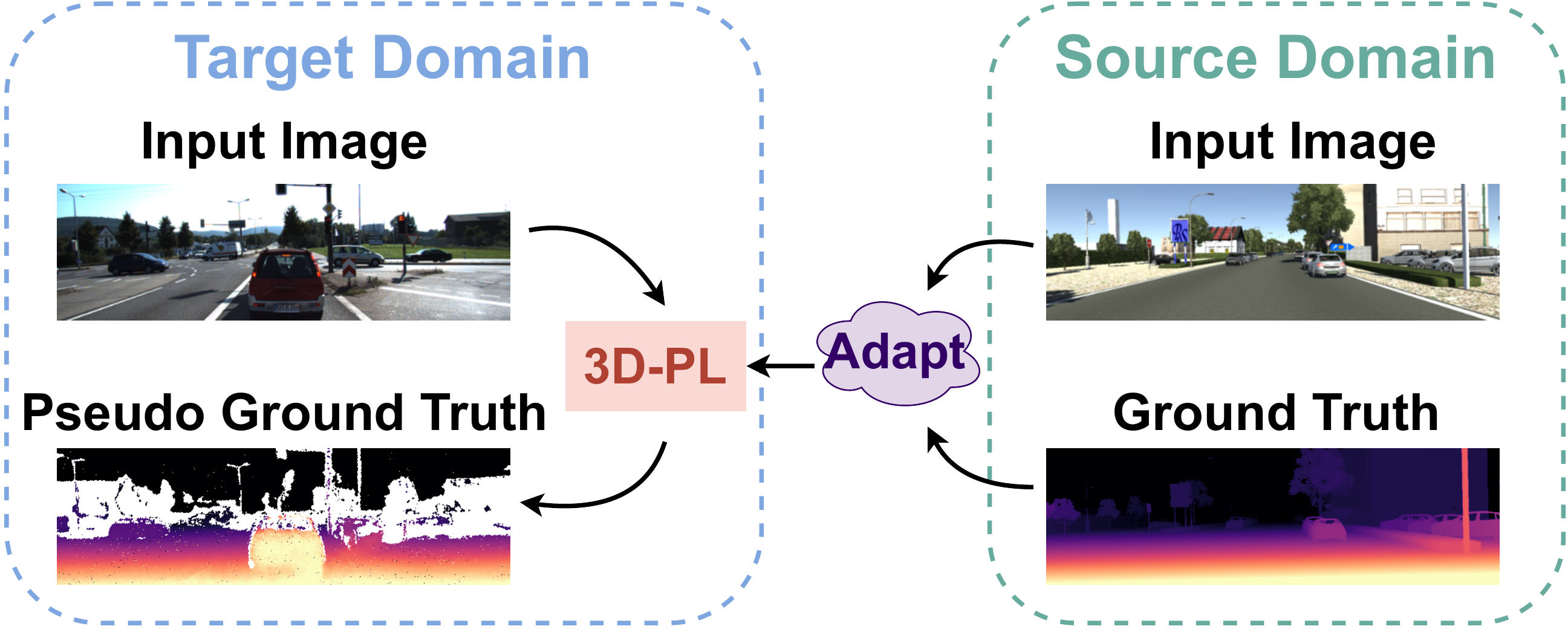}
         \caption{Overview of our proposed method for domain adaptation.}
         \label{fig.teaser.a}
     \end{subfigure}
     }
     \resizebox{0.42\linewidth}{!}{
     \begin{subfigure}{0.4\linewidth}
         \includegraphics[width=\linewidth]{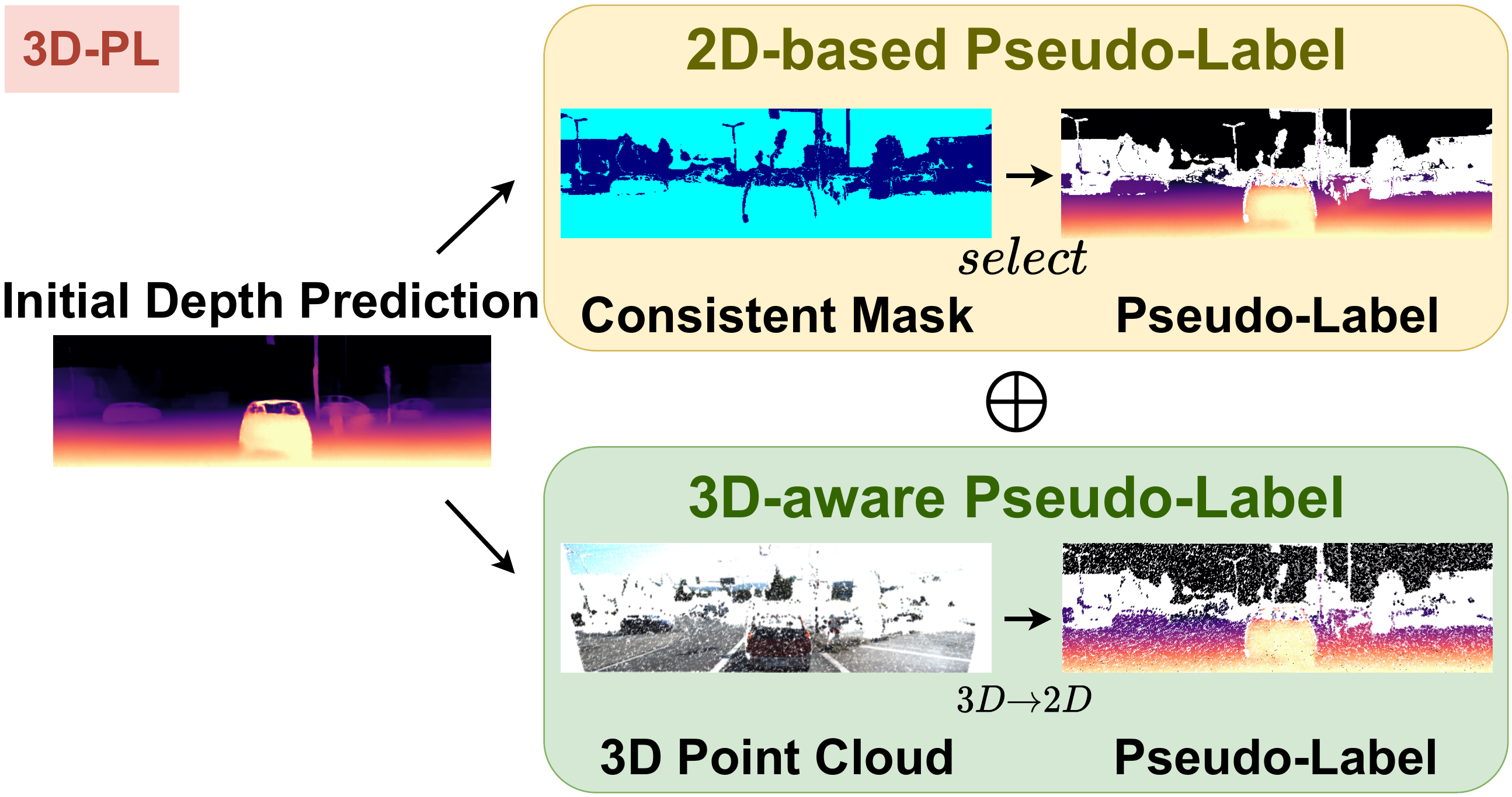}
         \caption{Basic concept behind our 3D-aware Pseudo-Labeling (3D-PL).}
         \label{fig.teaser.b}
     \end{subfigure}
     }
\centering
    \caption{\textbf{(a)} We propose a 3D-aware pseudo-labeling (3D-PL) technique to facilitate source-to-target domain adaptation for monocular depth estimation via pseudo-labeling on the target domain. \textbf{(b)} Our 3D-PL technique consists of 2D-based and 3D-aware pseudo-labels, where the former selects the pixels with highly-confident depth prediction (colorized by light blue in the consistent mask), while the latter performs 3D point cloud completion that provides refined pseudo-labels projected from 3D.}
    \label{fig:teaser}
\end{figure}

In the scenario of domain adaptation, two major techniques are commonly adopted to reduce the domain gap for depth estimation: 1) using adversarial loss \cite{kundu2018adadepth,chen2019crdoco, zheng2018t2net} for feature-level distribution alignment, or 2) leveraging style transfer between synthetic/real data to generate real-like images as pixel-level adaptation \cite{zheng2018t2net,zhao2019geometry}.
On the other hand, self-learning via pseudo-labeling the target real data is another powerful technique for domain adaptation~\cite{li2019bidirectional, zou2018unsupervised, vu2019advent}, yet less explored in the depth estimation task.
One reason is that, unlike tasks such as semantic segmentation that has the probabilistic output for classification to produce pseudo-labels, depth estimation is a regression task which requires specific designs for pseudo-label generation. In this paper, we propose novel pseudo-labeling methods in depth estimation for domain adaptation (see Fig. \ref{fig.teaser.a}).

To this end, we propose two mechanisms, 2D-based and 3D-aware methods, for generating pseudo depth labels (see Fig. \ref{fig.teaser.b}). For the 2D type, we consider the consistency of depth predictions when the model sees two images with the same content but different styles, i.e., the depth prediction can be more reliable for pixels with higher consistency. Specifically, we apply style transfer \cite{huang2017arbitrary} to the target real image and generate its synthetic-stylized version, and then find their highly-consistent areas in depth predictions as pseudo-labels.
However, this design may not be sufficient as it produces pseudo-labels only in certain confident pixels but ignore many other areas. Also, it does not take the fact that depth prediction is a 3D task into account.

To leverage the confident information obtained in our 2D-based pseudo-labeling process, we further propose to find the neighboring relationships in the 3D space via point cloud completion, so that our model is able to even select the pseudo-labels in areas that are not that confident, thus being complementary to 2D-based pseudo-labels.
Specifically, we first project 2D pseudo-labels to point clouds in the 3D space, and then utilize a 3D completion model to generate neighboring point clouds. Due to the help of more confident and accurate 2D pseudo-labels, it also facilitates 3D completion to synthesize better point clouds. Next, we project the completed point clouds back to depth values in the 2D image plane as our 3D-aware pseudo-labels. Since the 3D completion model learns the whole structural information in 3D space, it can produce reliable depth values that correct the original 2D pseudo-labels or expand extra pseudo-labels outside of the 2D ones.
We also note that, although pseudo-labeling for depth has been considered in the prior work \cite{lopez2020desc}, different from this work that needs a pre-trained panoptic segmentation model and can only generate pseudo-labels for object instances, our method does not have this limitation as we use the point cloud completion model trained on the source domain to infer reliable 3D-aware pseudo-labels on the target image.

We conduct extensive experiments by using the virtual KITTI dataset~\cite{gaidon2016virtual} as the source domain and the KITTI dataset~\cite{geiger2012we} as the real target domain.
We show that both of our 2D-based and 3D-aware pseudo-labeling strategies are complementary to each other and improve the depth estimation performance. In addition, following the stereo setting in GASDA~\cite{zhao2019geometry} where the stereo pairs are provided during training, our method can further improve the baselines and perform favorably against state-of-the-art approaches. Moreover, we directly evaluate our model on other unseen datasets, Make3D~\cite{saxena2008make3d}, and show good generalization ability against existing methods.
Here are our main contributions:
\begin{itemize}
\item We propose a framework for domain adaptive monocular depth estimation via pseudo-labeling, consisting of 2D-based and 3D-aware strategies that are complementary to each other.

\item We utilize the 2D consistency of depth predictions to obtain initial pseudo-labels, and then propose a 3D-aware method that adopts point cloud completion in the structural 3D space to refine and expand pseudo-labels.

\item We show that both of our 2D-based and 3D-aware methods have advantages against existing methods on several datasets, and when having stereo pairs during training, the performance can be further improved.

\end{itemize}
\section{Related Work}
\label{sec:related}
\noindent\textbf{Monocular Depth Estimation.}
Monocular depth estimation is to understand 3D depth information from a single 2D image. With the recent renaissance of deep learning techniques, supervised learning methods~\cite{eigen2014depth, liu2015learning, cao2017estimating, fu2018deep, yin2019enforcing, jiao2018look} have been proposed. Eigen \etal~\cite{eigen2014depth} first use a two-scale CNN-based network to directly regress on the depth, while Liu \etal~\cite{liu2015learning} utilize continuous CRF to improve depth estimation.
Furthermore, some methods propose different designs to extend the CNN-based network, such as changing the regression loss to classification~\cite{cao2017estimating, fu2018deep}, adding geometric constraints~\cite{yin2019enforcing}, and predicting with semantic segmentation~\cite{wang2015towards, jiao2018look}.

Despite having promising results, the cost of collecting image-depth pairs for supervised learning is expensive. Thus, several unsupervised~\cite{garg2016unsupervised, godard2017unsupervised, godard2019digging, mahjourian2018unsupervised, zhan2018unsupervised, tosi2019learning} or semi-supervised~\cite{guizilini2020robust, kuznietsov2017semi, amiri2019semi, ji2019semi} methods have been proposed to estimate disparity from the stereo pairs or videos. Garg \etal~\cite{garg2016unsupervised} warp the right image to reconstruct its corresponding left one (in a stereo pair) through the depth-aware geometry constraints, and take photometric error as the reconstruction penalty. Godard \etal~\cite{godard2017unsupervised} predict the left and right disparity separately, and enforce the left-right consistency to enhance the quality of predicted results. There are several follow-up methods to further improve the performance through semi-supervised manner~\cite{kuznietsov2017semi, amiri2019semi} and video self-supervision~\cite{godard2019digging, mahjourian2018unsupervised}.

\noindent\textbf{Domain Adaptation for Depth Estimation.}
{
Another way to tackle the difficulty of data collection for depth estimation is to leverage the domain adaptation techniques \cite{kundu2018adadepth, zheng2018t2net, chen2019crdoco, zhao2019geometry, lopez2020desc, pnvr2020sharingan}, where the synthetic data can provide full supervisions as the source domain and the real-world unlabeled data is the target domain.
Since depth estimation is a regression task, existing methods usually rely on style transfer/image translation for pixel-level adaptation \cite{atapour2018real}, adversarial learning for feature-level adaptation \cite{kundu2018adadepth}, or their combinations \cite{zheng2018t2net,zhao2019geometry}.
For instance, Atapour \etal~\cite{atapour2018real} transform the style of testing data from real to synthetic, and use it as the input to their depth prediction model that is only trained on the synthetic data. AdaDepth~\cite{kundu2018adadepth} aligns the distribution between the source and target domain at the latent feature space and the prediction level. T$^2$net~\cite{zheng2018t2net} further combines these two techniques, where they adopt both the synthetic-to-real translation network and the task network with feature alignment. They show that, training on the real stylized images brings promising improvement, but aligning features is not effective in the outdoor dataset. 

Other follow-up methods~\cite{chen2019crdoco, zhao2019geometry} take the bidirectional translation (real-to-synthetic and synthetic-to-real) and use the depth consistency loss on the prediction between the real and real-to-synthetic images. Moreover, some methods employ additional information to give constraints on the real image. GASDA~\cite{zhao2019geometry} utilizes stereo pairs and encourages the geometry consistency to align stereo images.
With a similar setting and geometry constraint to GASDA, SharinGAN~\cite{pnvr2020sharingan} maps both synthetic and real images to a shared image domain for depth estimation.
Moreover, DESC~\cite{lopez2020desc} adopts instance segmentation to apply pseudo-labeling using instance height and semantic segmentation to encourage the prediction consistency between two domains. Compared to these prior works, our proposed method provides direct supervisions on the real data in a simple and efficient pseudo-labeling way without any extra information.
}

\noindent\textbf{Pseudo-Labeling for Depth Estimation.}
In general, pseudo-labeling explores the knowledge learned from labeled data to infer pseudo ground truths for unlabeled data, which is commonly used in classification~\cite{lee2013pseudo, hu2021simple, taherkhani2021self, chen2019progressive, saito2017asymmetric} and scene understanding~\cite{zou2018unsupervised, zou2020pseudoseg, li2019bidirectional, pastore2021closer, chen2020digging, Zhao_UniDet_ECCV20, Paul_daseg_eccv20, mmtta_CVPR22} problems.
Only few depth estimation methods~\cite{lopez2020desc, yang2021self} adopt the concept of pseudo-labeling. DESC~\cite{lopez2020desc} designs a model to predict the instance height and converts the instance height to depth values as the pseudo-label for the depth prediction of the real image. Yang \etal \cite{yang2021self} generate the pseudo-label from multi-view images and design a few ways to refine pseudo-labels, including fusing point clouds from multi-views.
These methods succeed in producing pseudo-labels, but they require to have the instance information \cite{lopez2020desc} or multi-view images \cite{yang2021self}. Moreover, as \cite{yang2021self} is a multi-stereo task, it is easier to build a complete point cloud from multi views and render the depth map as pseudo-labels. Their task also focuses on the main object instead of the overall scene. In our method, we design the point cloud completion method to generate reliable 3D-aware pseudo-labels based on a single image that contains a real-world outdoor scene.

\begin{figure*}[t]
	\centering

    \resizebox{1\linewidth}{!}{
         \begin{subfigure}{1\textwidth}
         \centering
         \includegraphics[width=\textwidth]{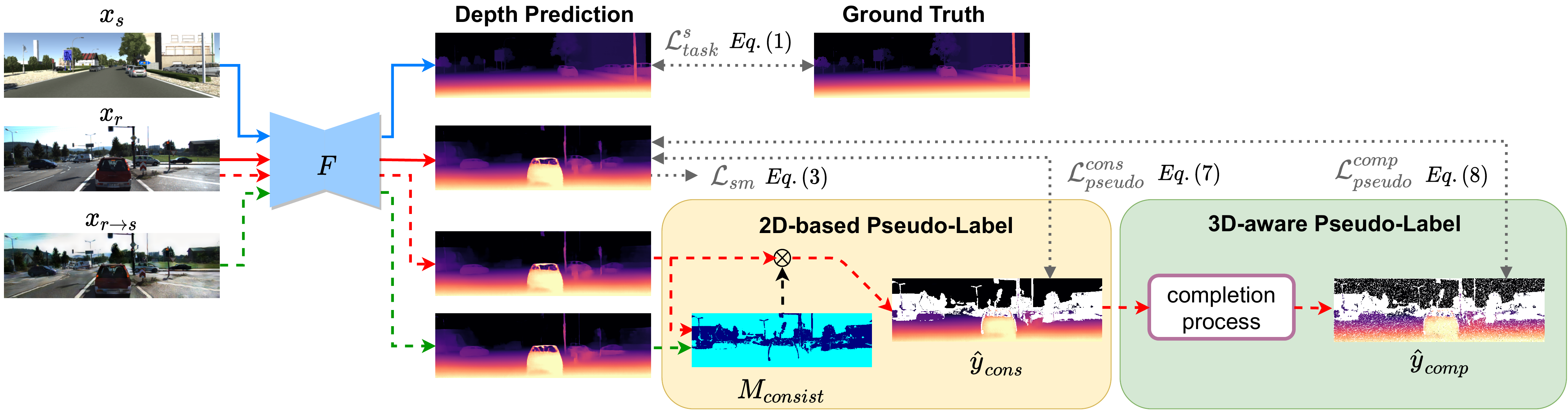}
         \caption{Our overall 3D-PL framework for pseudo-labeling.}
         \label{fig.sub.a}
        \end{subfigure}
         }
         
    \resizebox{1\linewidth}{!}{
          \begin{subfigure}{1\textwidth}
         \centering
         \includegraphics[width=\textwidth]{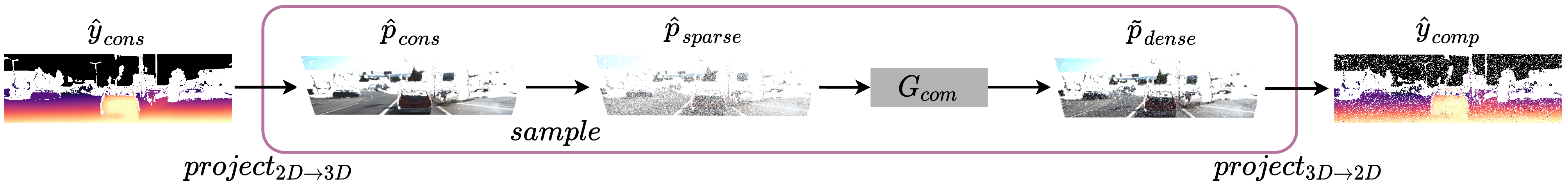}
         \caption{3D-aware pseudo-label generation via 3D completion.}
         \label{fig.sub.b}
     \end{subfigure}
             }

    \caption{\textbf{(a)} Illustration of our proposed 3D-PL framework together with the training objectives. $F$ is the depth prediction network, with input of the synthetic image $x_s$, the real image $x_r$, and the synthetic-stylized image $x_{r\rightarrow s}$. 
    In 3D-PL, we obtain 2D-based pseudo-labels $\hat{y}_{cons}$ through finding the region with consistent depth (light blue color in $M_{consist}$) across the predictions of $x_r$ and ${x}_{r\rightarrow s}$ (see Section~\ref{sec:2d_consistency_label}), while 3D-aware pseudo-labels $\hat{y}_{comp}$ are obtained via the 3D completion process.
    Here we denote solid lines as the computation flow where the gradients can be back-propagated, while the dashed lines indicate that pseudo-labels are generated offline based on the preliminary model in Section~\ref{sec:base}.
    \textbf{(b)} We first project the 2D-based pseudo-labels $\hat{y}_{cons}$ to the 3D point cloud $\hat{p}_{cons}$, followed by uniformly sub-sampling $\hat{p}_{cons}$ to sparse $\hat{p}_{sparse}$.
    Then, the completion network $G_{com}$ densifies $\hat{p}_{sparse}$ to obtain $\tilde{p}_{dense}$, in which we further project $\tilde{p}_{dense}$ back to 2D and produce 3D-aware pseudo-labels $\hat{y}_{comp}$ (see Section~\ref{sec:3d-aware-completion-label}).
    }
	\label{fig:model}
\end{figure*}

\section{Proposed Method}
\label{sec:method}
Our goal in this paper is to adapt the depth prediction model $F$ to the unlabeled real image $x_r$ as the target domain, where the synthetic image-depth pair $(x_s, y_s)$ in the source domain is provided for supervision. Without domain adaptation, the depth prediction model $F$ can be well trained on the synthetic data $(x_s, y_s)$, but it cannot directly perform well on the real image $x_r$ because of the domain shift. Thus, we propose our pseudo-labeling method to provide direct supervisions on target image $x_r$, which reduces the domain gap effectively.

Fig. \ref{fig:model} illustrates the overall pipeline of our method. To utilize our pseudo-labeling techniques, we first use the synthetic data to train a preliminary depth prediction model $F$, and then adopt this pretrained model to infer pseudo-labels on real data for self-training.
For pseudo-label generation, we propose 2D-based and 3D-aware schemes, where we name them as \textit{consistency label} and \textit{completion label}, respectively. We detail our model designs in the following sections.

\subsection{Preliminary Model Objectives}
\label{sec:base}
Here, we describe the preliminary objectives during our model pre-training by using the synthetic image-depth pairs $(x_s,y_s)$ and the real image $x_r$, including depth estimation loss and smoothness loss. Please note that, this is a common step before pseudo-labeling, in order to account for initially noisy predictions.

\noindent\textbf{Depth Estimation Loss.}
As synthetic image-depth pairs $(x_s,y_s)$ can provide the supervision, we directly
minimize the $L_1$ distance between the predicted depth $\tilde{y}_s = F(x_s)$ of the synthetic image $x_s$ and the ground truth depth $y_s$.
\begin{equation}
\label{equ:task_s_loss}
\mathcal{L}_{task}^{s}(F) = \lVert\tilde{y}_{s} - y_{s}\rVert_1.
\end{equation}
In addition to the synthetic images $x_s$, we follow the similar style translation strategy as~\cite{zheng2018t2net} to generate real-stylized images $x_{s\rightarrow r}$, in which $x_{s\rightarrow r}$ maintains the content of $x_s$ but has the style from a randomly chosen real image $x_r$. Note that, to keep the simplicity of our framework, we adopt the real-time style transfer AdaIN~\cite{huang2017arbitrary} (pretrained model provided by~\cite{huang2017arbitrary}) instead of training another translation network like \cite{zheng2018t2net}.
\begin{equation}
\label{equ:task_s2t_loss}
\mathcal{L}_{task}^{s\rightarrow r}(F) = \lVert\tilde{y}_{s\rightarrow r} - y_{s}\rVert_1.
\end{equation}

\noindent\textbf{Smoothness Loss.}
For the target image $x_r$, we adopt the smoothness loss as \cite{godard2017unsupervised,zheng2018t2net} to encourage the local depth prediction $\tilde{y}_r$ being smooth and consistent.
Since depth values are often discontinuous on the boundaries of objects, we weigh this loss with the edge-aware term:
\begin{equation}
\label{equ:sm_loss}
\mathcal{L}_{sm}(F)=e^{-\nabla x_{r}}\lVert\nabla \tilde{y}_{r}\rVert_1,
\end{equation}
where $\nabla$ is is the ﬁrst derivative along spatial directions.

\subsection{Pseudo-Label Generation}
With the preliminary loss functions introduced in Section \ref{sec:base} that pre-train the model, we then perform our pseudo-labeling process with two schemes.
First, 2D-based consistency label aims to find the highly confident pixels from depth predictions as pseudo-labels. Second, 3D-aware completion label utilizes a 3D completion model $G_{com}$ to refine some prior pseudo-labels and further extend the range of pseudo-labels (see Fig. \ref{fig:model}).

\subsubsection{2D-based Consistency Label}
\label{sec:2d_consistency_label}
A typical way to discover reliable pseudo-labels is to find confident ones, e.g., utilizing the softmax output from tasks like semantic segmentation~\cite{ li2019bidirectional}. However, due to the nature of the regression task in depth estimation, it is not trivial to obtain such 2D-based pseudo-labels from the network output. Therefore, we design a simple yet effective way to construct the confidence map via feeding the model two target images with the same content but different styles. Since pixels in two images have correspondence, our motivation is that, pixels that are more confident should have more consistent depth values across two predictions (i.e., finding pixels that are more domain invariant through the consistency of predictions from real images with different styles).

To achieve this, we first obtain the synthetic-stylized image $x_{r\rightarrow s}$ for the real image $x_r$, which combines the content of $x_r$ and the style of a synthetic image $x_s$, via AdaIN~\cite{huang2017arbitrary}. Then, we obtain depth predictions of these two images, $\tilde{y}_r = F(x_r)$, $\tilde{y}_{r\rightarrow s} = F(x_{r\rightarrow s})$, and calculate their difference. If the difference at one pixel is less then a threshold $\tau$, we consider this pixel as a more confident prediction to form the pseudo-label $\hat{y}_{cons}$.
The procedure is written as:
\begin{align}
M_{consist} = |\tilde{y}_r - \tilde{y}_{r\rightarrow s}| < \tau, \notag \\
\hat{y}_{cons} = M_{consist}\otimes \tilde{y}_r,
\label{equ:consistency_label}
\end{align}
where $M_{consist}$ is the binary mask for consistency, which records where pixels are consistent. $\tau$ is the threshold, set as $0.5$ in meter, and $\otimes$ is the element-wise product to filter the prediction $\tilde{y}_r$ of the target image.

\subsubsection{3D-aware Completion Label}
\label{sec:3d-aware-completion-label}
Since depth estimation is a 3D problem, we expand the prior 2D-based pseudo-label $\hat{y}_{cons}$ to obtain more pseudo-labels in the 3D space, so that the pseudo-labeling process can benefit from the learned 3D structure.
To this end, based on the 2D consistency label $\hat{y}_{cons}$, we propose a 3D completion process to reason neighboring relationships in 3D.
As shown in Fig. \ref{fig.sub.b}, the 3D completion process adopts the point cloud completion technique to learn from the 3D structure and generate neighboring points.

First, we project the 2D-based pseudo-label $\hat{y}_{cons}$ to point clouds $\hat{p}_{cons} = project_{2D \rightarrow 3D}(\hat{y}_{cons}) $ in the 3D space. In the projection procedure, we reconstruct each point $(x_i, y_i, z_i)$ from the image pixel $(u_i, v_i)$ with its depth value $d_i$ based on the standard pinhole camera model (more details and discussions are provided in the supplementary material).
Next, we uniformly sample points from $\hat{p}_{cons} $ to have sparse point clouds $\hat{p}_{sparse} = sample(\hat{p}_{cons}) $, followed by taking $\hat{p}_{sparse}$ as the input to the 3D completion model $G_{com}$ for synthesizing the missing points. Those generated points from the 3D completion model $G_{com}$ compose new dense point clouds $\tilde{p}_{dense} = G_{com}(\hat{p}_{sparse})$, and then we project each point $(\tilde{x_i}, \tilde{y_i}, \tilde{z_i})$ back to the original 2D plane as $(\tilde{u}_i, \tilde{v}_i)$ with updated depth value $\tilde{d}_i=\tilde{z}_i$.

\begin{figure*}[!t]
	\centering
    \includegraphics[width=1\textwidth]{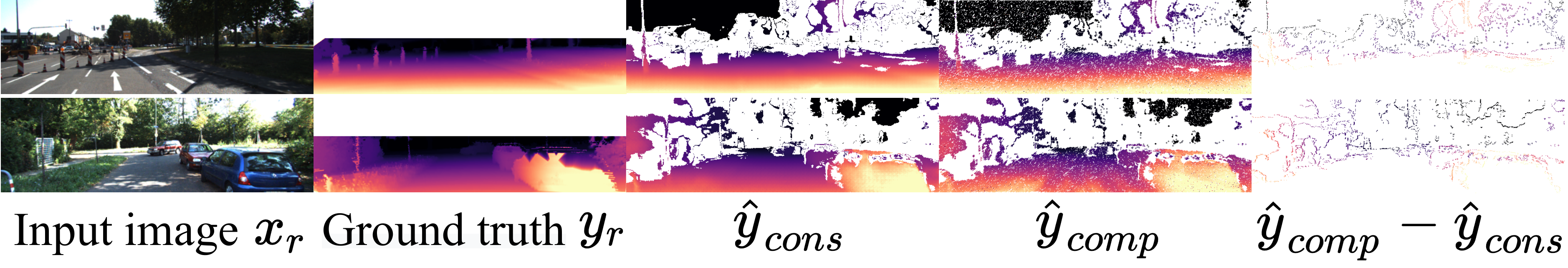}
    \caption{Examples for our pseudo-labels.
    The third and fourth columns are pseudo-labels for 2D-based $\hat{y}_{cons}$ and 3D-aware $\hat{y}_{comp}$. The final column represents the complementary pseudo-labels produced by $\hat{y}_{comp}$.
    Note that ground truth $y_r$ is the reference but not used in our model training.
    }
	\label{fig:pseudo-label}
\end{figure*}

Therefore, our 3D-aware pseudo-label $\hat{y}_{comp}$ (i.e., completion label) is formed by the updated depth value $\tilde{d}_i$.
Note that, as there could exist some projected points falling outside the image plane and not all the pixels on the image plane are covered by the projected points, we construct a mask $M_{valid}$ which records the pixels on the completion label $\hat{y}_{comp}$ where the projection succeeds, i.e., having valid $(\tilde{u}_i, \tilde{v}_i)$.
\begin{equation}
\label{equ:completion_label}
\hat{y}_{comp} = M_{valid}\otimes project_{3D\rightarrow 2D}(\tilde{p}_{dense}).
\end{equation}
In Fig. \ref{fig:pseudo-label}, we show that the 3D-aware completion label $\hat{y}_{comp}$ expands the pseudo-labels from the 2D-based consistency label $\hat{y}_{cons}$, i.e., visualizations in $\hat{y}_{comp} - \hat{y}_{cons}$ are additional pseudo-labels from the 3D completion process (please refer to Section~\ref{sec:effectiveness} for further analyzing the effectiveness of 3D-aware pseudo-labels).

\noindent\textbf{3D Completion Model.}
We pre-train the 3D completion model $G_{com}$ using the synthetic ground truth depth $y_s$ in advance and keep it fixed during our completion process. We project the entire $y_s$ to 3D point clouds $\hat p^{s} = project_{2D\rightarrow 3D}(y_s)$ and then perform the same process (i.e., sampling and completion) as in Fig. \ref{fig.sub.b} to obtain the generated dense point clouds $\tilde p_{dense}^{s}$. Since $\hat p^{s}$ is the ground truth point clouds of $\tilde p_{dense}^{s}$, we can directly minimize Chamfer Distance (CD) \cite{fan2017point} between these two point clouds to train the 3D completion model $G_{com}$, $\mathcal{L}_{cd}(G_{com})=CD(\hat p^{s}, \tilde p_{dense}^{s})$.

\subsection{Overall Training Pipeline and Objectives}
\label{subsec:train_strategy}
There are two training stages in our proposed method: the first stage is to train a preliminary depth model $F$, and the second stage is to apply the proposed pseudo-labeling techniques through this preliminary model. The loss in the first stage consists of the ones introduced in Section \ref{sec:base}:
\begin{equation}
\label{equ:total_loss_first}
\mathcal{L}_{base} = \lambda_{task} (\mathcal{L}_{task}^{s} + \mathcal{L}_{task}^{s\rightarrow r}) + \lambda_{sm}\mathcal{L}_{sm},
\end{equation} 
where $\lambda_{task}$ and $\lambda_{sm}$ are set as $100$ and $0.1$ respectively, following the similar settings in~\cite{zheng2018t2net}. Note that in our implementation, for every synthetic image $x_s$, we augment three corresponding real-stylized images $x_{s\rightarrow r}$, where their styles are obtained from three real images randomly drawn from the training set.

\noindent\textbf{Training with Pseudo-labels.}
In the second stage, we use our generated 2D-based and 3D-aware pseudo-labels in Eq.~\eqref{equ:consistency_label} and Eq.~\eqref{equ:completion_label} to provide direct supervisions on the target image $x_r$. Since the completion label $\hat{y}_{comp}$ is aware of the 3D structural information and can refine the prior 2D-based pseudo-labels $\hat{y}_{cons}$, we choose the completion label $\hat{y}_{comp}$ as the main reference if a pixel has both consistency label $\hat{y}_{cons}$ and completion label $\hat{y}_{comp}$. The 2D and 3D pseudo-label loss functions are respectively defined as:
\begin{align}
\mathcal{L}_{pseudo}^{cons}(F) = \lVert M_{valid}' \otimes (M_{consist} \otimes \tilde{y}_{r} - \hat{y}_{cons})\rVert_1, \\
\mathcal{L}_{pseudo}^{comp}(F) = \lVert M_{valid} \otimes \tilde{y}_{r} - \hat{y}_{comp}\rVert_1,
\end{align}
where $M_{valid}' = (1-M_{valid})$ is the inverse mask of $M_{valid}$. In addition to the two pseudo-labeling objectives, we also include the supervised synthetic data to maintain the training stability. The total objective of the second stage is:
\begin{equation}
\begin{aligned}
\label{equ:total_loss_second}
\mathcal{L}_{total} = \alpha(\lambda_{cons}\mathcal{L}_{pseudo}^{cons} + \lambda_{comp}\mathcal{L}_{pseudo}^{comp}) \\
+ (1-\alpha) \lambda_{task}^{s}\mathcal{L}_{task}^{s} + \lambda_{sm}\mathcal{L}_{sm},
\end{aligned}
\end{equation} 
where $\alpha$ set as $0.7$ is the proportion ratio between the supervised loss of the synthetic and real image. $\lambda_{task}^{s}$, $\lambda_{cons}$, $\lambda_{comp}$, and $\lambda_{sm}$ are set as $100$, $ 1$, $0.1$, and $0.1$, respectively. Here we do not include the $\mathcal{L}_{task}^{s\rightarrow r}$ loss as in Eq. \eqref{equ:total_loss_first} to make the model training more focused on the real-domain data.

\noindent\textbf{Stereo Setting.}
The training strategy mentioned above is under the condition that we can only access the monocular single image of the real data $x_r$. In addition, if the stereo pairs are available during training as the setting in GASDA~\cite{zhao2019geometry}, we can further include the geometry consistency loss $\mathcal{L}_{tgc}$ in \cite{zhao2019geometry} to our proposed method (more details are in the supplementary material):
\begin{equation}
\begin{aligned}
\label{equ:total_loss_stereo}
\mathcal{L}_{stereo} = \mathcal{L}_{total} + \lambda_{tgc}\mathcal{L}_{tgc},
\end{aligned}
\end{equation} 
where $\mathcal{L}_{total}$ is the loss in Eq.~\eqref{equ:total_loss_second}, and $\lambda_{tgc}$ is set as $50$ following \cite{zhao2019geometry}.

\section{Experimental Results}
\label{sec:experiments}
In summary, we conduct experiments for the synthetic-to-real benchmark when only single images or stereo pairs are available during training. Then we show ablation studies to demonstrate the effectiveness of the proposed pseudo-labeling methods. Moreover, we provide discussion to validate the effectiveness of our 3D-aware pseudo-labeling method. Finally, we directly evaluate our models on two real-world datasets to show the generalization ability.  More results and analysis are provided in the supplementary material.

\noindent\textbf{Datasets and Evaluation Metrics.}
We adopt Virtual KITTI (vKITTI)~\cite{gaidon2016virtual} and real KITTI~\cite{geiger2012we} as our source and target datasets respectively. vKITTI contains $21,260$ synthetic image-depth pairs of the urban scene under different weather conditions.
Since the maximum depth ground truth values are different in vKITTI and KITTI, we clip the maximum value to $80m$ as~\cite{zheng2018t2net}.
For evaluating the generalization ability, we use the KITTI Stereo~\cite{menze2015object} and Make3D~\cite{saxena2008make3d} datasets following the prior work \cite{zhao2019geometry}.
We use the same depth evaluation metrics as~\cite{zheng2018t2net, zhao2019geometry}, including four types of errors and three types of accuracy metrics.

\noindent\textbf{Implementation Details.}
Our depth prediction model $F$ adopts the same U-net~\cite{ronneberger2015u} structure as~\cite{zheng2018t2net}. Following~\cite{xiang20203ddepthnet}, the 3D completion model $G_{com}$ is modified from PCN~\cite{yuan2018pcn} with PointNet~\cite{qi2017pointnet}.
We implement our model based on the Pytorch framework with NVIDIA Geforce GTX 2080 Ti GPU. All networks are trained with the Adam optimizer. The depth prediction model $F$ and 3D completion model $G_{com}$ are trained from scratch with learning rate $10^{-4}$ and linear decay after 10 epochs. We train $F$ for 20 epochs in the first stage and 10 epochs in the second stage, and pre-train $G_{com}$ for 20 epochs.
The style transfer network AdaIN is pre-trained without any finetuning. 

\begin{table}[!t]

\renewcommand{\footnotesize}{\fontsize{5pt}{7pt}\selectfont}
\footnotesize

\centering
\renewcommand{\arraystretch}{1.1}

\setlength{\tabcolsep}{1.1pt}

\captionsetup{font={small}}
\caption{Quantitative results on KITTI in the single-image setting, where we denote the best results in bold. 
For the training data, ``K'', ``CS'', and ``S'' indicate KITTI~\cite{geiger2012we}, CityScapes~\cite{cordts2016cityscapes}, and virtual-KITTI~\cite{gaidon2016virtual} datasets respectively. We highlight the rows in gray for those methods using the domain adaptation (DA) techniques.
}

\begin{tabular}{c||c|c|c||cccc|ccc}
\hline
\multirow{2}{*}{Method} & \multirow{2}{*}{Supervised} & \multirow{2}{*}{Dataset} & \multirow{2}{*}{Cap} & \multicolumn{4}{c|}{Error Metrics (lower, better)} & \multicolumn{3}{c}{Accuracy Metrics (higher, better)} \\ \cline{5-11}
                  &                   &                   &                   &  Abs Rel   &  Sq Rel   &  RMSE   & RMSE log   &   $\;\;\;\delta<1.25$    &   $\delta<1.25^2$    &  $\delta<1.25^3$    \\
\hline\hline
  Eigen {\it et al.}~\cite{eigen2014depth}                &     Yes              &        K           &     $80m$              &  0.203   &  1.548   &  6.307   &  0.282  &   0.702    &   0.890    &  0.958    \\
  Liu {\it et al.}~\cite{liu2015learning}                &      Yes             &      K             &  $80m$                  &  0.202   &  1.614   &  6.523   & 0.275   &  0.678     &  0.895     &  0.965    \\
     Zhou {\it et al.}~\cite{zhou2017unsupervised}             &    No               &    K            &       $80m$            & 0.208    &  1.768   &  6.856   & 0.283   &  0.678     &   0.885    &  0.957    \\
     Zhou {\it et al.}~\cite{zhou2017unsupervised}              &   No                &    K+CS               &     $80m$              & 0.198    &  1.836   &  6.565   &  0.275  &   0.718    &   0.901    &  0.960    \\
\hline
All synthetic      &   No             &  S            &   $80m$             &  0.253   &  2.303   &   6.953   & 0.328   & 0.635    &  0.856    & 0.937 \\
All real&  No             &  K     &   $80m$             &  0.158   &  1.151   & 5.285   &  0.238   & 0.811    & 0.934    &  0.970\\
\hline
\cellcolor[HTML]{EFEFEF} AdaDepth~\cite{kundu2018adadepth}              &     \cellcolor[HTML]{EFEFEF} No             &     \cellcolor[HTML]{EFEFEF}  K+S(DA)            &   \cellcolor[HTML]{EFEFEF}   $80m$             & \cellcolor[HTML]{EFEFEF} 0.214   & \cellcolor[HTML]{EFEFEF} 1.932   & \cellcolor[HTML]{EFEFEF} 7.157   &\cellcolor[HTML]{EFEFEF}0.295   & \cellcolor[HTML]{EFEFEF}  0.665    & \cellcolor[HTML]{EFEFEF}  0.882    & \cellcolor[HTML]{EFEFEF} 0.950 \\
   \cellcolor[HTML]{EFEFEF}    T$^2$Net~\cite{zheng2018t2net}              &  \cellcolor[HTML]{EFEFEF}    No             &  \cellcolor[HTML]{EFEFEF}     K+S(DA)            &  \cellcolor[HTML]{EFEFEF}    $80m$               &\cellcolor[HTML]{EFEFEF} 0.182     &\cellcolor[HTML]{EFEFEF}    1.611     &\cellcolor[HTML]{EFEFEF}    6.216     &\cellcolor[HTML]{EFEFEF}    0.265     &\cellcolor[HTML]{EFEFEF}    0.749     &\cellcolor[HTML]{EFEFEF}    0.898     &\cellcolor[HTML]{EFEFEF}   0.959 \\

   \cellcolor[HTML]{EFEFEF}    3D-PL (Ours)           &  \cellcolor[HTML]{EFEFEF}    No             &  \cellcolor[HTML]{EFEFEF}     K+S(DA)            &  \cellcolor[HTML]{EFEFEF}    $80m$            &   \cellcolor[HTML]{EFEFEF} \bf{0.169}     &\cellcolor[HTML]{EFEFEF}        \bf{1.371}     &\cellcolor[HTML]{EFEFEF}        \bf{6.037}    &\cellcolor[HTML]{EFEFEF}      \bf{0.256}     &\cellcolor[HTML]{EFEFEF}      \bf{0.759}     &\cellcolor[HTML]{EFEFEF}     \bf{0.904}     &\cellcolor[HTML]{EFEFEF}     \bf{0.961}  \\

\hline \hline

     Garg {\it et al.}~\cite{garg2016unsupervised} & No & K & $50m$ & 0.169 & 1.080 & 5.104 & 0.273 & 0.740 & 0.904 & 0.962 \\
\hline
All synthetic      & No      &      S       &   $50m$             &  0.244   &  1.771   &  5.354   &  0.313   &   0.647    &  0.866    & 0.943 \\
     All real&    No       &     K           &    $50m$             & 0.151   &  0.856   &  4.043   & 0.227   & 0.824    &  0.940    &  0.973 \\
\hline
  \cellcolor[HTML]{EFEFEF}  AdaDepth~\cite{kundu2018adadepth}              & \cellcolor[HTML]{EFEFEF}     No             &    \cellcolor[HTML]{EFEFEF}   K+S(DA)            &  \cellcolor[HTML]{EFEFEF}    $50m$             & \cellcolor[HTML]{EFEFEF} 0.203   & \cellcolor[HTML]{EFEFEF} 1.734   & \cellcolor[HTML]{EFEFEF} 6.251   & \cellcolor[HTML]{EFEFEF} 0.284   & \cellcolor[HTML]{EFEFEF}  0.687    & \cellcolor[HTML]{EFEFEF}  0.899    & \cellcolor[HTML]{EFEFEF} 0.958 \\

  \cellcolor[HTML]{EFEFEF}    T$^2$Net~\cite{zheng2018t2net}              &  \cellcolor[HTML]{EFEFEF}    No             &  \cellcolor[HTML]{EFEFEF}     K+S(DA)            &  \cellcolor[HTML]{EFEFEF}    $50m$             & \cellcolor[HTML]{EFEFEF} 0.168   & \cellcolor[HTML]{EFEFEF} 1.199   & \cellcolor[HTML]{EFEFEF}  4.674   & \cellcolor[HTML]{EFEFEF} 0.243   &  \cellcolor[HTML]{EFEFEF} 0.772    &  \cellcolor[HTML]{EFEFEF} 0.912    &  \cellcolor[HTML]{EFEFEF} 0.966 \\

  \cellcolor[HTML]{EFEFEF}    3D-PL (Ours)             &  \cellcolor[HTML]{EFEFEF}    No             &  \cellcolor[HTML]{EFEFEF}     K+S(DA)            &  \cellcolor[HTML]{EFEFEF}    $50m$            &   \cellcolor[HTML]{EFEFEF} \bf{0.162}    &\cellcolor[HTML]{EFEFEF}        \bf{1.049}     &\cellcolor[HTML]{EFEFEF}        \bf{4.463}   &\cellcolor[HTML]{EFEFEF}    
  \bf{ 0.239}   &\cellcolor[HTML]{EFEFEF}      \bf{0.776}     &\cellcolor[HTML]{EFEFEF}     \bf{0.916}     &\cellcolor[HTML]{EFEFEF}      \bf{0.968}\\

\hline
\end{tabular}
\label{tb:single_qua}
\end{table}

\begin{figure*}[!t]
	\centering
	{
    \includegraphics[width=0.95\textwidth]{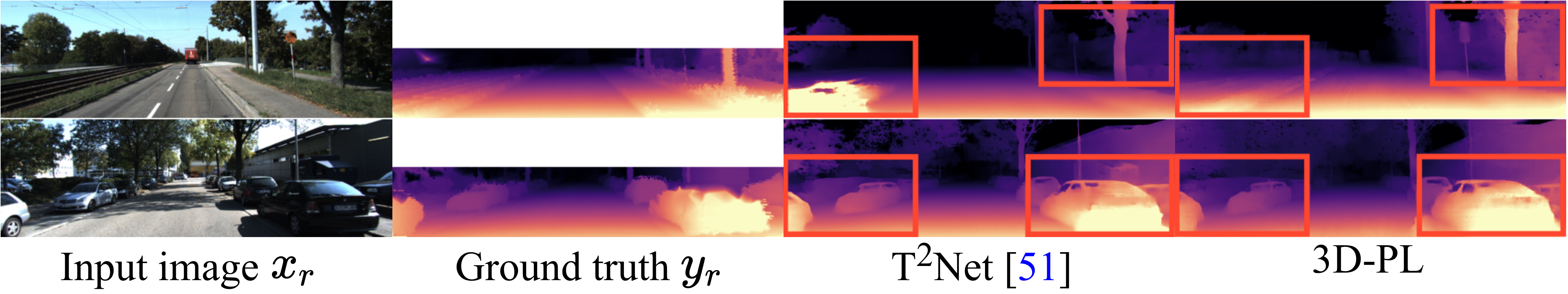}}
    \caption{Qualitative results on KITTI in the single-image setting.
    We show that our 3D-PL produces more accurate results for the tree and grass (upper row) and better shapes in the car (bottom row), compared to the T$^2$Net~\cite{zheng2018t2net} method.
    }
	\label{fig:qualitative_no_stereo}
\end{figure*}

\subsection{Synthetic-to-Real Benchmark}
We follow~\cite{zheng2018t2net} to use $22,600$ KITTI images from $32$ scenes as the real training data, and evaluate the performance on the eigen test split~\cite{eigen2014depth} of $697$ images from other 29 scenes. Following~\cite{zhao2019geometry}, we evaluate the depth prediction results with the ground truth depth less than $80m$ or $50m$. There are two real-data training settings in domain adaptation for monocular depth estimation: 1) only single real images are available and we cannot access binocular or semantic information as~\cite{zheng2018t2net}; 2) stereo pairs are available during training, so that geometry consistency can be leveraged as~\cite{zhao2019geometry}. Our pseudo-labeling method does not have an assumption of the data requirement, and hence we conduct experiments in these two different data settings as mentioned in Section \ref{subsec:train_strategy}.

\noindent\textbf{Single-image Setting.}
In this setting, we can only access monocular real images in the whole training process, as the overall objective in Eq.~\eqref{equ:total_loss_second}. 
Table \ref{tb:single_qua} shows the quantitative results, where the domain adaptation methods are highlighted in gray. ``All synthetic/All real'' are only trained on synthetic/real image-depth pairs, which can be viewed as the lower/upper bound. Our 3D-PL method outperforms T$^2$Net (state-of-the-art) in every metric, especially $13\%$ and $15\%$ improvement in the ``Sq Rel'' error of $50m$ and $80m$.
Fig. \ref{fig:qualitative_no_stereo} shows the qualitative results, where we compare our 3D-PL with T$^2$Net~\cite{zheng2018t2net}. In the upper row, 3D-PL produces more accurate results for the tree and grass, while T$^2$Net predicts too far and close respectively. In the lower row, our result has a better shape in the right car and more precise depth for the left two cars.

\begin{table*}[!t]
\renewcommand{\footnotesize}{\fontsize{5pt}{7pt}\selectfont}
\footnotesize
\footnotesize
\centering
\renewcommand{\arraystretch}{1.1}
\setlength{\tabcolsep}{1.1pt}
\captionsetup{font={small}}
\caption{Quantitative results on KITTI with having stereo pairs during training.
}
\begin{tabular}{p{4cm}||c||cccc|ccc}
\hline
\multirow{2}{*}{Method} & \multirow{2}{*}{Cap} & \multicolumn{4}{c|}{Error Metrics (lower, better)} & \multicolumn{3}{c}{Accuracy Metrics (higher, better)} \\ \cline{3-9}
     &                &  Abs Rel   &  Sq Rel   &  RMSE   & RMSE log   &   $\delta<1.25$    &   $\delta<1.25^2$    &  $\delta<1.25^3$    \\
\hline\hline
  Synthetic + Stereo    &   $80m$  &0.151&     1.176&     5.496&     0.237&     0.787&     0.926&     0.972    \\
  T$^2$Net~\cite{zheng2018t2net}  + Stereo    &   $80m$  &0.154&     1.115&     5.504&     0.233&     0.800&     0.929&     0.971    \\
  GASDA~\cite{zhao2019geometry} (Stereo)   &   $80m$  & 0.149   &  1.003   &  4.995   & 0.227   &      0.824    &   0.941    & 0.973    \\
  DESC~\cite{lopez2020desc} + Stereo &   $80m$  &  0.122&    0.946&    5.019&    0.217&    0.843&    0.942&    0.974    \\
  SharinGAN~\cite{pnvr2020sharingan} (Stereo)    &   $80m$  &   0.116&      0.939&       5.068&      0.203&      0.850&      0.948&      0.978    \\
\hline
 3D-PL + Stereo&   $80m$  &      \bf{0.113}&     \bf{0.903}&    \bf{ 4.902}&     \bf{0.201}&     \bf{0.859}&     \bf{0.952}&    \bf{ 0.979}\\
\hline \hline
  Synthetic + Stereo    &   $50m$ &     0.145&     0.909&     4.204&     0.224&     0.800&     0.934&     0.975       \\
  T$^2$Net~\cite{zheng2018t2net}  + Stereo    &   $50m$  &0.148&     0.828&     4.123&     0.219&     0.815&     0.938&     0.975     \\
  GASDA~\cite{zhao2019geometry} (Stereo)   &   $50m$  &  0.143   &   0.756   &   3.846  &   0.217   &     0.836    &    0.946    &   0.976\\
  DESC~\cite{lopez2020desc} + Stereo &   $50m$  &   0.116&    0.725&    3.880&    0.206&    0.855&    0.948&    0.976\\
SharinGAN~\cite{pnvr2020sharingan} (Stereo) &   $50m$  &   0.109&      0.673&       3.770&        0.190&      0.864&      0.954&       0.981    \\
\hline
 3D-PL + Stereo&   $50m$  &   \bf{ 0.106}&     \bf{0.641}&    \bf{ 3.643}&    \bf{ 0.189}&   \bf{  0.872}&    \bf{ 0.958}&     \bf{0.982}\\
\hline
\end{tabular}
\label{tb:stereo_qua}
\end{table*}

\noindent\textbf{Stereo-pair Setting.}
If stereo pairs are available, we can utilize the geometry constraints to have self-supervised stereo supervisions as~\cite{zhao2019geometry} using the objective in Eq.~(\ref{equ:total_loss_stereo}).
Table \ref{tb:stereo_qua} shows that our 3D-PL achieves the best performance among state-of-the-art methods. In particular, without utilizing any other clues from real-world semantic annotation, 3D-PL outperforms DESC~\cite{lopez2020desc} with $12\%$ lower ``Sq Rel'' error in the stereo scenario.
This shows that our pseudo-labeling is able to generate more reliable pseudo-labels over the single-image setting.

\begin{figure*}[!t]
\centering
    \includegraphics[width=0.95\textwidth]{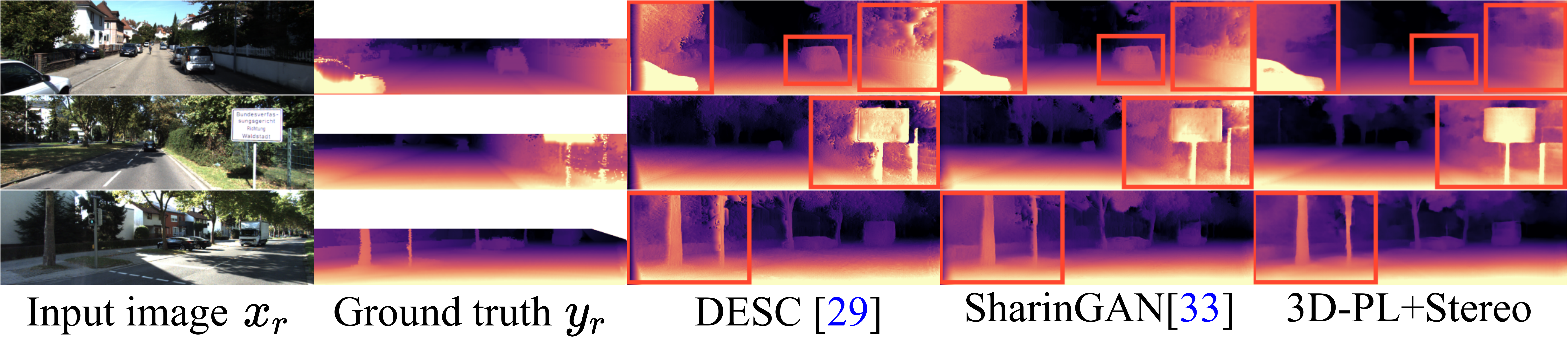}
    \caption{Qualitative results on KITTI with having stereo pairs during training.
    We show that our 3D-PL produces better results on the overall structure (e.g., tree, wall, and car in the top row), closer objects (e.g., road sign in the middle row, tree in the bottom row), and shapes (e.g., traffic  light  in  the  bottom  row), compared to DESC~\cite{lopez2020desc} and SharinGAN~\cite{pnvr2020sharingan}.
    }
\label{fig:qualitative_stereo}
\end{figure*}

Fig. \ref{fig:qualitative_stereo} shows qualitative results, where we compare our 3D-PL with DESC~\cite{lopez2020desc} + Stereo and SharinGAN~\cite{pnvr2020sharingan}. 3D-PL produces better results on the overall structure (e.g., tree, wall, and car in the top row). For challenging situations such as closer objects standing alone and hiding in a complicated farther background (e.g., road sign in the middle row, tree in the bottom row), other methods tend to produce similar depth values as the background, while 3D-PL predicts better object shape and distinguish the object from the background even if it is very thin. (e.g., traffic light in the bottom row). This shows the benefits of our 3D-aware pseudo-labeling design, which reasons the 3D structural information.

\begin{table}[!t]
\scriptsize
\centering
\renewcommand{\arraystretch}{1.1}
\setlength{\tabcolsep}{1.5pt}
\captionsetup{font={small}}
\caption{Ablation study on KITTI in the single-image setting. 
}
\begin{tabular}{p{3.5cm}|cccc}
\hline
   \multirow{2}{*}{Method} & \multicolumn{4}{c}{Error Metrics (lower, better)} \\ \cline{2-5}
                    &  Abs Rel   &  Sq Rel   &  RMSE   & RMSE log\\
\hline \hline
  Only synthetic   &  0.244   &  1.771   &  5.354   &  0.313     \\
   $+\hat{y}_{cons}$ (all pixels) &     0.166&     1.125&     4.557&     0.244\\
   $+\hat{y}_{cons}$ (confident) &  0.163&    1.095&   4.555&    0.243\\
   $+ \hat{y}_{comp}$(confident)   &  0.164&     1.054&     4.473&     \bf{0.239}\\

  \hline
  3D-PL ($\hat{y}_{comp}$ all pixels)   &          \bf{ 0.161}&      1.070&     4.504&     0.240\\
  3D-PL ($\hat{y}_{comp}$ confident)     &     0.162&     \bf{ 1.049}&    \bf{  4.463}&    \bf{  0.239 }\\
\hline
\end{tabular}
\label{tb:ablation}
\end{table}


\subsection{Ablation Study}
\label{subsec:ablation}
{
We demonstrate the contributions of our model designs in Table \ref{tb:ablation} using the ``$50m$ Cap'' and single-image settings, where ``Only synthetic'' trains only on the supervised synthetic image-depth pairs.

\noindent\textbf{Importance of Pseudo-labels.}
First, we show that either using the 2D-based or 3D-aware pseudo-labels improve the performance, i.e., ``$+\hat{y}_{cons}$ (confident)'' and `` $+\hat{y}_{comp}$ (confident)''.
Then, our final model in ``3D-PL ($\hat{y}_{comp}$ confident)'' further improves depth estimation, and shows the complementary properties of using both 2D-based and 3D-aware pseudo-labels.

\noindent\textbf{Importance of Consistency Mask.}
We show the importance of having the consistency mask in Eq. \eqref{equ:consistency_label} as the confidence measure.
For the 2D-based pseudo-label, we compare the result of using the consistency mask ``$+\hat{y}_{cons}$ (confident)'' and the one using the entire depth prediction as the pseudo-label, ``$+\hat{y}_{cons}$ (all pixels)''. With the consistency mask, it has $3\%$ lower in the ``Sq Rel'' error.
Moreover, this consistency mask also improves 3D-aware pseudo-labeling when we project depth values to point clouds for 3D completion.
When inputting all the pixels for this process, i.e., ``3D-PL ($\hat{y}_{comp}$ all pixels)'', this may include less accurate depth values for performing 3D completion, which results in less reliable pseudo-labels compared to our final model using the confident pixels, i.e., ``3D-PL ($\hat{y}_{comp}$ confident)''.

}

\subsection{Effectiveness of 3D-aware Pseudo-labels}\label{sec:effectiveness}

To show the impact of 3D-aware pseudo-labels, we compute the proportions of pixels chosen as 2D/3D pseudo-labels in each image and take the average as the final statistics. 
The effectiveness of 3D-aware pseudo-labels is in two-fold: \textbf{refine} and \textbf{extend} from 2D-based pseudo-labels. In Table \ref{tb:proportion}, the initial proportion of confident 2D-based pseudo-labels ``2D only ($+\hat{y}_{cons}$)'' is 48.91\% among image pixels. As stated in Section~\ref{subsec:train_strategy}, 3D-PL improves original 2D labels, which results in 43.63\% refined and 3.9\% extended labels. The rightmost subfigure of Fig.~\ref{fig:pseudo-label} visualizes extended labels $\hat{y}_{comp} - \hat{y}_{cons}$, in which it shows that the improved performance is contributed by the larger proportion of 3D-aware pseudo-labels.

\noindent\textbf{Ability of pseudo-label refinement.}
Since the 2D-based and 3D-aware pseudo-labels may have the duplication on the same pixel, we conduct experiments to use either $\hat{y}_{cons}$ or $\hat{y}_{comp}$ as the reference when such cases happen.
In Table \ref{tb:choose}, choosing $\hat{y}_{comp}$ as the main reference has the better performance, which indicates that updating the pseudo-label of a pixel from original $\hat{y}_{cons}$ to $\hat{y}_{comp}$ brings the positive effect. This validates that $\hat{y}_{comp}$ can refine the prior 2D-based pseudo-labels since it is aware of the 3D structural information.


\begin{table}[!t]
\scriptsize
\centering
\renewcommand{\arraystretch}{1.1}
\setlength{\tabcolsep}{2pt}
\captionsetup{font={small}}
\caption{Statistics of pixel proportion in our 2D/3D pseudo-labels. ``R'' and ``E'' indicate ``refined'' and ``extended''.}
\begin{tabular}{c|c|c}
\hline
 Method   & 2D Proportion & 3D Proportion \\
\hline
 2D only ($+\hat{y}_{cons}$)   &     48.91\% & 0\% \\
 3D-PL & 5.28\% & 43.63\% (R) + 3.9\% (E) \\
\hline
\end{tabular}
\label{tb:proportion}
\end{table}

\begin{table}[!t]
\scriptsize
\centering
\renewcommand{\arraystretch}{1.1}
\setlength{\tabcolsep}{4pt}
\captionsetup{font={small}}
\caption{Results of using either the 2D-based $\hat{y}_{cons}$ or the 3D-aware $\hat{y}_{comp}$ pseudo-label as the reference, when there is a duplication on both pseudo-labels.
}
\begin{tabular}{p{3.5cm}|cccc}
\hline
   \multirow{2}{*}{\quad\quad\;  Main Reference} & \multicolumn{4}{c}{Error Metrics (lower, better)} \\ \cline{2-5}
                    &  Abs Rel   &  Sq Rel   &  RMSE   & RMSE log\\
\hline \hline
   Completion Label $\hat{y}_{comp}$  &  \bf{0.162}&    \bf{ 1.049}&     \bf{4.463}&     \bf{0.239}    \\
   Consistency Label  $\hat{y}_{cons}$  & 0.164&     1.095&     4.529&     0.243\\
\hline
\end{tabular}
\label{tb:choose}
\end{table}

\subsection{Generalization to Real-world Datasets}
{
\noindent\textbf{KITTI Stereo.}
We evaluate our model on 200 images of KITTI stereo 2015~\cite{menze2015object}, which is a small subset of KITTI images but has different ways of collecting groundtruth of depth information. Since the ground truth of KITTI stereo has been optimized for the moving objects, it is denser than LiDAR, especially for the vehicles. Note that, this benefits DESC~\cite{lopez2020desc} in this evaluation as their method relies on the instance information from the pre-trained segmentation model. Table \ref{tb:kitti_stereo} shows the quantitative results, where our 3D-PL in both single-image and stereo settings performs competitively against existing methods.

}

\begin{table*}[!t]
\renewcommand{\footnotesize}{\fontsize{5pt}{7pt}\selectfont}
\footnotesize
\centering
\renewcommand{\arraystretch}{1.1}
\setlength{\tabcolsep}{1.1pt}
\captionsetup{font={small}}
\caption{Quantitative results on KITTI stereo 2015 benchmark~\cite{menze2015object}. ``S$^\diamond$'' denotes synthetic data that \cite{atapour2018real} captures from GTA~\cite{richter2016playing}. ``Supervised'' represents whether the method is trained on KITTI stereo.
}
\begin{tabular}{c||c|c||cccc|ccc}
\hline
\multirow{2}{*}{Method} & \multirow{2}{*}{Supervised} & \multirow{2}{*}{Dataset} & \multicolumn{4}{c|}{Error Metrics (lower, better)} & \multicolumn{3}{c}{Accuracy Metrics (higher, better)} \\ \cline{4-10}
                  &                   &                &  Abs Rel   &  Sq Rel   &  RMSE   & RMSE log   &   $\delta<1.25$    &   $\delta<1.25^2$    &  $\delta<1.25^3$    \\
\hline\hline
      Godard {\it et al.}~\cite{godard2017unsupervised}             &     No              &     K               &  0.124   &  1.388   &  6.125   &  0.217  &  0.841     &  0.936     & 0.975     \\
     Godard {\it et al.}~\cite{godard2017unsupervised}             &     No              &     K+CS             &  0.104   &  1.070   &  5.417   &  0.188  &  0.875     &  0.956     & 0.983     \\
     Atapour {\it et al.}~\cite{atapour2018real}            &      No             &       K+S$^\diamond$(DA)    &  0.101   &  1.048   &  5.308   & 0.184   &   0.903    &   0.988    &  0.992 \\
     \hline
    T$^2$Net ~\cite{zheng2018t2net} &   No  &    K+S(DA)   &     0.155&     1.731&     6.510&     0.237&     \bf{0.800} &     \bf{0.921} &     \bf{0.969}  \\
    3D-PL &      No             &  K+S(DA)   & \bf{0.147} &     \bf{1.352} &     \bf{6.157} &     \bf{0.233} &     \bf{0.800} &     0.918&     0.967\\
   
     \hline\hline
     GASDA~\cite{zhao2019geometry} (Stereo) &      No             &       K+S(DA)            &  0.106   &   0.987   &   5.215   &  0.176   &   0.885    &   0.963    & 0.986 \\
     DESC~\cite{lopez2020desc} + Stereo &      No             &  K+S(DA)  &\bf{0.085}&     \bf{0.781}&     4.490&     0.158&     0.909&     0.967&     0.986 \\
     SharinGAN~\cite{pnvr2020sharingan} (Stereo) &      No             &  K+S(DA)  &0.092&      0.904&      4.614&      0.159&      0.906&      0.969&      0.987  \\
     3D-PL + Stereo &      No             &  K+S(DA)   &\bf{0.085}&     0.830&    \bf{ 4.489}&    \bf{ 0.149}&     \bf{0.915}&    \bf{ 0.971}&    \bf{ 0.988} \\ 

\hline
\end{tabular}
\label{tb:kitti_stereo}
\end{table*}

\begin{table}[!t]
\scriptsize
\centering
\renewcommand{\arraystretch}{1.1}
\setlength{\tabcolsep}{1.1pt}
\captionsetup{font={small}}
\caption{Quantitative results on Make3D~\cite{saxena2008make3d}. ``Supervised'' represents whether the method is trained on Make3D.
}
\begin{tabular}{c||c||ccc}
\hline
\multirow{2}{*}{Method} & \multirow{2}{*}{${\rm Supervised}$} & \multicolumn{3}{c}{Error Metrics (lower, better)} \\ \cline{3-5}
                  &                           &  Abs Rel   &  Sq Rel   &  RMSE  \\
\hline\hline
     Karsch {\it et al.}~\cite{karsch2014depth}     &   Yes  & 0.398  & 4.723  & 7.801  \\
     Laina {\it et al.}~\cite{laina2016deeper}      &   Yes  & 0.198  & 1.665  & 5.461  \\
     AdaDepth~\cite{kundu2018adadepth}    &   Yes  & 0.452  & 5.71   & 9.559  \\
\hline
     Godard {\it et al.}~\cite{godard2017unsupervised}   &   No  &   0.505   &  10.172   &  10.936   \\
     AdaDepth~\cite{kundu2018adadepth}   &   No  &   0.647   & 12.341   &  11.567    \\
    T$^2$Net ~\cite{zheng2018t2net} &   No  &   0.508   &  6.589   &  8.935    \\
     Atapour {\it et al.}~\cite{atapour2018real}     &      No   &    0.423 &  9.343   &   9.002  \\
     GASDA~\cite{zhao2019geometry} &      No             &      0.403          &  6.709   &  10.424  \\
    DESC~\cite{lopez2020desc} &      No             &0.393&     4.604&     8.126    \\
    SharinGAN~\cite{pnvr2020sharingan} &      No             &      0.377          &  4.900  &  8.388  \\
    S2R-DepthNet~\cite{chen2021s2r} & No & 0.490&  10.681& 10.892\\
\hline
     3D-PL  &      No & \bf{0.352}&    \bf{3.539}&    \bf{7.967} \\
\hline
\end{tabular}

\label{tab:make3d}
\end{table}

\noindent\textbf{Make3D Dataset.}
{ 
Moreover, we directly evaluate the model on the Make3D dataset~\cite{saxena2008make3d} without any ﬁnetuning. We choose $134$ test images with central image crop and clamp the depth value to $70m$, following \cite{godard2017unsupervised}.
Here, since Make3D is a different domain from the KITTI training data, we apply the single-image model to reduce the strong domain-related constraints such as the stereo supervisions.
In Table \ref{tab:make3d}, 3D-PL achieves the best performance compared to other approaches.
It is also worth mentioning that 3D-PL outperforms  \textcolor{black}{the domain generalization method~\cite{chen2021s2r} and supervised method~\cite{karsch2014depth} by $66\%$ and $25\%$ in ``Sq Rel''}, showing the promising generalization capability.
}

\section{Conclusions}
\label{sec:conclusion}
In this paper, we introduce a domain adaptation method for monocular depth estimation. We propose 2D-based and 3D-aware pseudo-labeling mechanisms, which utilize knowledge from synthetic domain as well as 3D structural information to generate reliable pseudo depth labels for real data. Extensive experiments show that our pseudo-labeling strategies are able to improve depth estimation in various settings against several state-of-the-art domain adaptation approaches, as well as achieving good performance in unseen datasets for generalization.

\noindent\textbf{Acknowledgement.}
This project is supported by MOST (Ministry of Science and Technology, Taiwan) 111-2636-E-A49-003 and 111-2628-E-A49-018-MY4.

%
%

\bibliographystyle{splncs04}
\bibliography{main}

\begin{center}
\textbf{\Large Supplemental Materials}
\end{center}

\section{Stereo Setting}
In this section, we provide the details for objective and the overall training pipeline in the stereo-pair setting as introduced in Section 3.3 of the main paper.
\label{sec:stereo}
\subsection{Objectives}
When stereo pairs are available, we utilize the geometry constraints to have self-supervised stereo supervisions as GASDA~\cite{zhao2019geometry} using the geometry consistency loss. The stereo pairs contain the left image $x_r$, which is also used in the single-image setting, and the corresponding right image. Here we denote the left and right real images as $x^{left}, x^{right}$ (we ignore the notation $r$ in the real image like $x^{left}_r$ for simplicity) and their depth prediction $\tilde{y}^{left} = F(x^{left}), \tilde{y}^{right} = F(x^{right})$. The geometry consistency loss in GASDA~\cite{zhao2019geometry} is a reconstruction penalty between the real left image $x^{left}$ and the warped left image $\tilde{x}^{left}$. 
\begin{equation}
\label{equ:tgc_left}
\mathcal{L}_{tgc}^{left}(F)=\eta\frac{1-SSIM(x^{left}, \tilde{x}^{left})}{2}+\mu||x^{left}-\tilde{x}^{left}||,
\end{equation}
where $\eta$ and $\mu$ are set as $0.85$ and $0.15$ respectively following ~\cite{zhao2019geometry}.
The warped left image $\tilde{x}^{left}$ is obtained from the disparity $a$ and the right image $x^{right}$ with bilinear sampling~\cite{jaderberg2015spatial} following~\cite{godard2017unsupervised}:
\begin{equation}
\label{equ:warp_left}
\tilde{x}^{left} = x^{right} - a, 
\end{equation}
Since we know the camera parameters when collecting the stereo images, we can convert the depth prediction $\tilde{y}^{left}$ of left image to the disparity $a$ through:
\begin{equation}
\label{equ:disparity_left}
a = \frac{b\cdot f}{\tilde{y}^{left}} ,
\end{equation}
where $b$ is the baseline distance between the two cameras and $f$ is the focal length, both parameters are known in the stereo-pair setting. In addition to reconstructing $\tilde{x}^{left}$ from the right image $x^{right}$, we also warp $\tilde{y}^{right}$ and $x^{left}$ to get $\tilde{x}^{right}$ in our experiments using a similar process and loss $\mathcal{L}_{tgc}^{right}$.
Finally, our geometry consistency loss is $\mathcal{L}_{tgc} = \mathcal{L}_{tgc}^{left} + \mathcal{L}_{tgc}^{right}$.
%
\subsection{Overall Training Pipeline}
In our stereo-pair setting, there are also two training stages for training a preliminary depth model $F$ and applying the proposed pseudo-labeling techniques through this preliminary model.

\noindent\textbf{Training a preliminary depth model $F$.} 
In the stereo-pair setting, we follow the single-image setting to use Eq.(6) in the main paper and train a preliminary depth model $F$ for $20$ epochs and further train another $10$ epochs with adding $\mathcal{L}_{tgc}$:
\begin{equation}
\label{equ:base_stereo}
\mathcal{L}_{base}^{stereo} = \lambda_{task}\mathcal{L}_{task}^{s} +  \lambda_{sm}\mathcal{L}_{sm} + \lambda_{tgc}\mathcal{L}_{tgc},
\end{equation}
where $\lambda_{task}$, $\lambda_{sm}$, and $\lambda_{tgc}$ are set as $100$, $0.1$, and $50$ respectively. Here we do not include the $\mathcal{L}_{task}^{s\rightarrow r}$ loss to make the model training more focused on the real-domain data. In the second stage, the overall loss is defined as Eq.(10) in the main paper.
\section{Sensitivity Analysis}
\label{sec:sensitivity}

{
In this section, we analyze the impact of different parameters, such as threshold or the weights in the loss. All experiments are performed in the single-image setting. 
}

\subsection{Threshold $\tau$}
{
Table \ref{tb:threshold} shows our results under different threshold $\tau$ as defined in Eq.(4) of the main paper, which controls the range of pseudo-label. The higher threshold means more pseudo-labels are chosen but may not be accurate, while the lower one can obtain more precise pseudo-labels but the amount is less. As shown in Table \ref{tb:threshold}, our method performs robustly under a reasonable range of $\tau$ (e.g., 0.3 to 1 meter).
}

\begin{table}[htp]
\footnotesize
\centering
\renewcommand{\arraystretch}{1.1}
\setlength{\tabcolsep}{4pt}
\begin{tabular}{p{3.5cm}|cccc}
\hline
   \multirow{2}{*}{\quad\quad\; Threshold} & \multicolumn{4}{c}{Error Metrics (lower, better)} \\ \cline{2-5}
                    &  Abs Rel   &  Sq Rel   &  RMSE   & RMSE log\\
\hline \hline
  \quad \quad\quad $\tau = 0.1 $   & \bf{ 0.161}&     1.048&     4.497&   \bf{   0.239 } \\
  \quad \quad\quad $\tau = 0.3 $   &     0.162&   \bf{   1.045}&     4.476&  \bf{    0.239}\\
  \quad \quad\quad \underline{$\tau = 0.5 $}  &  0.162&     1.049&    \bf{  4.463}&    \bf{  0.239 }\\
  \quad \quad\quad $\tau = 1  $  & \bf{  0.161}&     1.053&  \bf{    4.463}&   \bf{   0.239}   \\
  \quad \quad\quad $\tau = 2  $  & \bf{  0.161}&     1.060&     4.468&   \bf{   0.239}\\
  \quad \quad\quad $\tau = 3  $  &0.162&     1.066&     4.473&    \bf{  0.239}\\

\hline
\end{tabular}
\captionsetup{font={small}}
\caption{Our results of different thresholds $\tau$. The unit of $\tau$ is meter. Underline denotes our final setting.}
\label{tb:threshold}
\end{table}

\subsection{Proportion of Pseudo-label Loss $\alpha$}
\label{subsec:sensitivity_alpha}
{Table \ref{tb:alpha_peudo} shows the experiments of using different weight proportion between the pseudo-label loss on real data and the task loss on synthetic data, where $\alpha$ is defined in Eq.(9) of the main paper.
With increasing the weight of pseudo-label loss, e.g., $\alpha=0.3$ to $0.7$, the performance is gradually improved, which shows the benefits of our proposed pseudo-labeling strategy. However, the performance drops when $\alpha$ becomes too large, which indicates the importance of having the accurate supervisions from the synthetic data to stabilize model training.
}

\begin{table}[htp]
\footnotesize
\centering
\renewcommand{\arraystretch}{1.1}
\setlength{\tabcolsep}{4pt}
\begin{tabular}{p{1.2cm}|p{1.2cm}|cccc}
\hline
   \multirow{2}{*}{$\alpha$} & \multirow{2}{*}{$1-\alpha$} & \multicolumn{4}{c}{Error Metrics (lower, better)} \\ \cline{3-6}
               &     &  Abs Rel   &  Sq Rel   &  RMSE   & RMSE log\\
\hline \hline
  $0.3$ & $0.7$    &    \bf{ 0.161}&     1.051&     4.520&     0.241\\
  $0.5$ & $0.5$  & \bf{0.161}&     1.053&     4.490&     0.240  \\
  \underline{$0.7$}&\underline{ $0.3$ }  &  0.162&     1.049&     \bf{4.463}&    \bf{ 0.239}\\
  $0.9$ & $0.1$    &     0.164&    \bf{ 1.043}&     4.508&     0.240\\
  $1$ & $0$  &0.191&     1.175&     4.472&     0.253 \\
\hline
\end{tabular}
\captionsetup{font={small}}
\caption{Our results of using different proportions $\alpha$ between the pseudo-label loss ($\alpha$) and the task loss ($1-\alpha$). Underline denotes our final setting.}
\label{tb:alpha_peudo}
\vspace{-5mm}
\end{table}

\subsection{Weighted Terms $\lambda_{cons}$ and $\lambda_{comp}$}
{
Table \ref{tb:lambda_pseudo} shows the results of different values of weighted terms  ($\lambda_{cons}$, $\lambda_{comp}$) between 2D-based and 3D-aware pseudo-label loss($\mathcal{L}_{pseudo}^{cons}$, $\mathcal{L}_{pseudo}^{comp}$), defined in Eq.(9) of the main paper.
As shown in Table \ref{tb:lambda_pseudo}, our method performs robustly under a reasonable range of $\lambda_{cons}$ and $\lambda_{comp}$ if they do not become too large.
We also note that, since the 2D position projected from 3D has a little scale shift to the original 2D pixel on the image plane, there exists scale difference between $\mathcal{L}_{pseudo}^{cons}$($\approx 10^{-3}$) and $\mathcal{L}_{pseudo}^{comp}$($\approx 10^{-2}$). Thus, we use $10$ times $\lambda_{cons}$ than $\lambda_{comp}$ as our final setting.
}

\begin{table}[htp]
\footnotesize
\centering
\renewcommand{\arraystretch}{1.1}
\setlength{\tabcolsep}{4pt}
\begin{tabular}{p{1.2cm}|p{1.2cm}|cccc}
\hline
   \multirow{2}{*}{$\lambda_{cons}$} & \multirow{2}{*}{$\lambda_{comp}$} & \multicolumn{4}{c}{Error Metrics (lower, better)} \\ \cline{3-6}
               &     &  Abs Rel   &  Sq Rel   &  RMSE   & RMSE log\\
\hline \hline
$0.1$ & $0.01$   &  \bf{ 0.162}&     1.062&     4.711&     0.247   \\
$0.1$ & $0.1$   & 0.163&    \bf{ 1.045}&    4.483&     0.240  \\
  \underline{$1$} & \underline{$0.1$} & \bf{ 0.16}2&     1.049&    \bf{ 4.463}&    \bf{ 0.239 }\\
  $1$ & $1$   &0.169&     1.097&    \bf{ 4.463}&     0.243\\
  $10$ & $1$    &   0.168&     1.102&     4.485&     0.243  \\
$10$ & $10$   & 0.171&     1.138&     4.504&     0.244\\
\hline
\end{tabular}
\captionsetup{font={small}}
\caption{Our results of using different values of weighted term ($\lambda_{cons}, \lambda_{comp}$) between 2D and 3D pseudo-label loss. Underline denotes our final setting.}
\label{tb:lambda_pseudo}
\end{table}

\section{More Experiments}
In this section, we provide experiments for showing the effectiveness of 3D completion model $G_{com}$, the comparison with 2D depth completion model, the design choices of the depth estimation loss $\mathcal{L}_{task}$, and the model complexity.
\label{sec:experiments_supp}

\subsection{Effectiveness of $G_{com}$}
We verify whether the 3D completion model $G_{com}$ is well trained. To this end, 
we simply take one sequence ``0018'' out of Virtual KITTI dataset as the testing set while the remaining is the training set, and then use the same training procedure stated in the main paper to train our 3D completion model $G_{com}$.
%
During evaluation, we first project the 2D ground truth depth $y_s$ in testing set to 3D point clouds and uniformly sample them to have sparse point cloud $\hat p_{sparse}^{s}$, and then we take $\hat p_{sparse}^{s}$ as the input of $G_{com}$ to obtain the result of completion $\tilde{p}_{dense}^{s}$. Finally, we project $\tilde p_{dense}^{s}$ to the 2D depth map $\tilde{y}_{dense}^{s}$ and measure the depth accuracy with its original ground truth $y_s$. Table \ref{tb:g_com} shows that $G_{com}$ has the ability to produce precise and reasonable 3D completion results.


\begin{table}[htp]
\footnotesize
\centering
\renewcommand{\arraystretch}{1.1}
\setlength{\tabcolsep}{4pt}
\begin{tabular}{p{4cm}|ccc}
\hline
   \multirow{2}{*}{\quad\quad\;  Method} & \multicolumn{3}{c}{Accuracy Metrics (higher, better)} \\ \cline{2-4}
                    &  $\delta<1.25$    &   $\delta<1.25^2$    &  $\delta<1.25^3$\\
\hline \hline
   3D completion model $G_{com}$  &  0.976&    0.991&    0.995    \\
\hline
\end{tabular}
\captionsetup{font={small}}
\caption{Performance of the completion model $G_{com}$ trained on the synthetic dataset.}
\vspace{-5mm}
\label{tb:g_com}
\end{table}

We also provide details for network architecture and sampling strategy of our completion model $G_{com}$.

\noindent \textbf{Network Architecture of $G_{com}$.}
 3D completion model $G_{com}$ is modified from PCN~\cite{yuan2018pcn}. We follows~\cite{xiang20203ddepthnet} to adjust the PCN network, including the encoder and the decoder. The encoder of our completion model $G_{com}$ is simplified to one PointNet\cite{qi2017pointnet} layer. Our decoder only uses the second stage of point generation in PCN, and we take our sparse point cloud as the ``Coarse Output'' in PCN. The whole network architecture will be made available to the public.

\noindent \textbf{Sampling Strategy.} We ``uniformly'' sample 3D point cloud into 30720 (25\% of pixel number in an image) sparse points as the input to the 3D completion model $G_{com}$. The 3D point cloud before sampling is projected from 2D depth map through the projection mechanism introduced in Section~\ref{subsec:projection_mechanism}. During the pre-training process of $G_{com}$, we sample the point cloud projected from synthetic ground truth depth $y_s$ to sparse point cloud $\hat{p}_{sparse}^s$. In 3D-aware  pseudo-labeling generation, we project 2D pseudo-labels to 3D as point clouds $\hat{p}_{cons}$ and sample $\hat{p}_{cons}$ to sparse point cloud $\hat{p}_{sparse}$.

\vspace{2mm}
\subsection{Comparison with 2D Depth Completion Model}
{
Since there exists 2D depth completion methods which are also able to complete depth values directly on 2D depth-maps/image~\cite{eldesokey2020uncertainty, park2020non, ma2019self}, we compare our 3D point cloud completion model $G_{com}$ with a 2D depth completion model~\cite{eldesokey2020uncertainty} to validate the necessity of our 3D-aware approach. We apply a recent 2D depth completion model~\cite{eldesokey2020uncertainty} to the sparse depth map sampled from our confident area with two types of training setting. One is pre-trained model provided by the author, and the other one is the model trained from scratch on vKITTI~\cite{gaidon2016virtual} with the same setting as our completion model $G_{com}$. 

Note that our 3D completion model $G_{com}$ is only trained on vKITTI~\cite{gaidon2016virtual} and the 2D depth completion model provided by the author is pre-trained on KITTI~\cite{geiger2012we}, so the 2D depth completion model accesses more information from the real domain. We replace our 3D completion model $G_{com}$ with the 2D depth completion model~\cite{eldesokey2020uncertainty} to generate pseudo-labels for training depth prediction model $F$. As shown in Table~\ref{tb:psuedo_label_com}, even ``+ 2D depth completion~\cite{eldesokey2020uncertainty}(pre-trained by authors)'' is pre-trained on KITTI supervisedly (i.e., using the ground truth depths for training), our proposed 3D-aware approach (i.e., ``+ 3D-aware completion label $\hat{y}_{cons}$'') provides better performance in all metrics.
In addition, we re-train the 2D depth completion model~\cite{eldesokey2020uncertainty} with the same training setting as ours (i.e., trained on vKITTI), and our proposed 3D-aware approach reaches $29\%$ lower error on the ``Sq Rel'' metric.
This shows that our 3D completion model $G_{com}$, which explicitly considers the 3D structural information, is able to produce more reliable pseudo-labels than the 2D depth completion models.
}


\begin{table}[htp]
\renewcommand{\footnotesize}{\fontsize{8pt}{10pt}\selectfont}
\footnotesize
\centering
\renewcommand{\arraystretch}{1.3}
\setlength{\tabcolsep}{4pt}
\begin{tabular}{p{7cm}|cccc}
\hline
Method & Abs Rel   &  Sq Rel   &  RMSE   & RMSE log\\
\hline
  + 2D depth completion~\cite{eldesokey2020uncertainty} (pre-trained by authors)   &\bf{0.164}&     1.068&     4.746&     0.247\\
  + 2D depth completion~\cite{eldesokey2020uncertainty} (trained on vKITTI)   &0.186&     1.476&     6.125&     0.310\\
 + 3D-aware completion label $\hat{y}_{cons}$ (Ours)   &    \bf{0.164}&     \bf{1.054}&     \bf{4.473}&     \bf{0.239} \\
  
\hline
\end{tabular}
\captionsetup{font={small}}
\caption{Training depth prediction model $F$ by using the pseudo-labels generated from different completion models. Note that ``+ 2D depth completion~\cite{eldesokey2020uncertainty} (pre-trained by authors)'' is pre-trained on KITTI~\cite{geiger2012we}, in which the 2D depth completion model~\cite{eldesokey2020uncertainty} has the supervision on depth directly from the real domain, while our model is trained on vKITTI.}
\label{tb:psuedo_label_com}
\end{table}

\subsection{Design Choices of Depth Estimation Loss $\mathcal{L}_{task}$}
As stated in Section~\ref{subsec:sensitivity_alpha}, when training with pseudo-labels, it is important to have accurate supervisions on the depth estimation loss $\mathcal{L}_{task}$ to stabilize model training. In Eq.(9) of the main paper, we retain $\mathcal{L}_{task}^{s}$ as the depth estimation to make the model training more focused on the real-domain data. While there exists another option $\mathcal{L}_{task}^{s\rightarrow r}$ for the depth estimation loss, as $\mathcal{L}_{task}^{s\rightarrow r}$ considers real-stylized images, images produced by style transfer may not align with their original depth ground truths well. Table~\ref{tb:de_loss} shows the experiments of adopting different options for the depth estimation loss in Eq.(9) of the main paper, which demonstrates that using $\mathcal{L}_{task}^{s}$ instead of $\mathcal{L}_{task}^{s\rightarrow r}$ has lower errors.

\begin{table}[htp]
\footnotesize
\centering
\renewcommand{\arraystretch}{1.1}
\setlength{\tabcolsep}{4pt}
\begin{tabular}{p{4cm}|cccc}
\hline
Method &  Abs Rel   &  Sq Rel   &  RMSE   & RMSE log\\
\hline

 using both $\mathcal{L}_{task}^{s}$ and $\mathcal{L}_{task}^{s\rightarrow r}$   & \bf{0.160}&     1.074&     4.611&     0.246\\
 using $\mathcal{L}_{task}^{s\rightarrow r}$ only  & 0.161&     1.090&     4.635&     0.247 \\
 using $\mathcal{L}_{task}^{s}$ only  &     0.162&     \bf{1.049}&     \bf{4.463}&    \bf{ 0.239}\\
  
\hline
\end{tabular}
\captionsetup{font={small}}
\caption{Different options for the depth estimation loss $\mathcal{L}_{task}$ in Eq.(9) of the main paper.}
\label{tb:de_loss}
\vspace{-5mm}
\end{table}

\subsection{Model Complexity}
We analyze the model complexity by computing the number of parameters and the training/testing time for our models. We have depth prediction model $F$ and the completion model $G_{com}$. The completion model $G_{com}$ only contains 1.324M parameters, which is much smaller than the depth model $F$ (54.565M). 
The training time for depth model $F$ and completion model $G_{com}$ are 80 and 21 hours. During testing, only the depth model $F$ is required, where it does not introduce additional overheads compared to normal model inference (0.014 seconds for a 192$\times$640 image as used in ~\cite{zhao2019geometry, zheng2018t2net}).

\subsection{Application on 3D Object Detection}
\begin{figure*}[h]
	\centering
	{
    \includegraphics[width=1\textwidth]{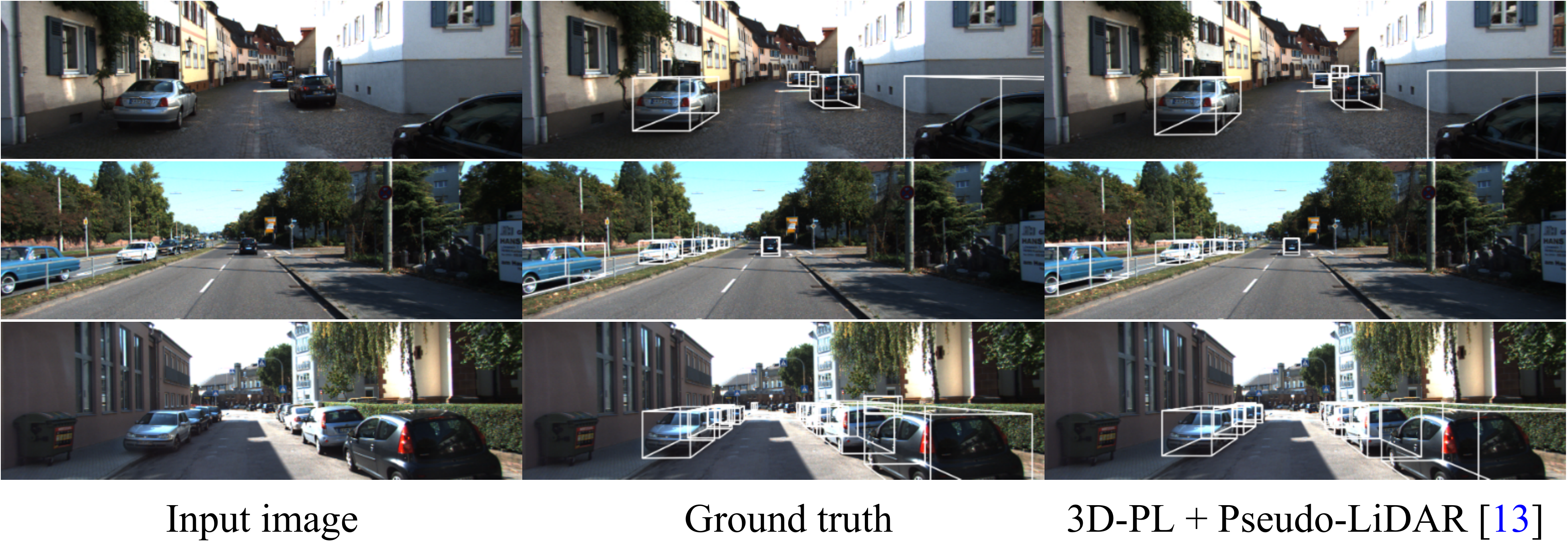}}
    \caption{3D object detection results using 3D-PL. 
    } 
	\label{fig:3d_object_detection}
\end{figure*}

To show the effectiveness of our depth result, we apply the final depth prediction to the 3D object detection task. We adopt Pseudo-LiDAR~\cite{wang2019pseudo} to convert our generated depth map to pseudo LiDAR, and take the pseudo LiDAR as the input to the 3D object detection model. We show example results in Figure~\ref{fig:3d_object_detection} compared to the ground truths. We also follow \cite{wang2019pseudo} to evaluate the result on the validation set of KITTI object detection benchmark for the ``car'' category. With the IoU threshold at 0.7, the average precision for the 3D object box detection (AP$_{3D}$) is 15.8\%, 12.3\%, and 11.2\% for easy, moderate, and hard cases, respectively. 

\vspace{5mm}
\section{Details for 2D/3D Projection}
\label{sec:projection}
We provide the implementation details for the projection procedure between 2D and 3D, including the projection mechanism and some discussions.
\subsection{Projection Mechanism}
\label{subsec:projection_mechanism}
{
\noindent\textbf{Projection from 2D to 3D.}
We aim to reconstruct each point $(x_i, y_i, z_i)$ in the 3D space from the 2D image pixel $(u_i, v_i)$ with its depth value $d_i$ based on the standard pinhole camera model. We assume the size of image is $H \times W$ and the pixel positions on the original image plane are ${\{(u_i, v_i)\}}^{H\times W}_{i=1}$, where each pixel $(u_i, v_i)$ has the corresponding depth value $d_i$.
Then, we project the point from 2D to 3D through $project_{2D\rightarrow 3D}$ to obtain 3D point $(x_i, y_i, z_i)$ in the 3D point cloud $\hat{y}_{cons}$:
\begin{equation}
\label{equ:project2to3}
x_i=\frac{d_i^* (u_i-o_x)}{f}, y_i=\frac{d_i^*(v_i-o_y)}{f}, z_i=d_i^*,
\end{equation}
where $f$ is the focal length, $o_x$ and $o_y$ are the 2D position of camera center, $d_i^* = d_i + \varepsilon$, $\varepsilon$ is a shift to convert relative depth value $d_i$ to the absolute depth value from the camera center.
Please note that, a single image has infinite possible 3D reconstruction depending on different camera parameters. Since our objective of the 3D completion model is to learn the structure and the depth relationship in the 3D space, we do not need to restore exactly the same setting as the image being captured in the real world. On the other hand, as we cannot know the camera parameters of the real data, we hence set up reasonable projection parameters on our own and use the same setting in training the 3D completion model and finding 3D-aware pseudo-labels. In experiments, we adopt the same focal length $f$ as virtual KITTI~\cite{gaidon2016virtual} and set $\varepsilon$ as $40$. Normally, $\varepsilon$ is set equal to the focal length $f$, but such setting would lead to large values for $x_i$ and $y_i$ coordinates as indicated in Eq.~\eqref{equ:project2to3}.
We therefore in experiments adopt the normalized depth values and fix the focal length to seek for a suitable shift $\varepsilon$, which gives a reasonable scale of 3D coordinates and still maintains the relationship between depth values.
}

{
\noindent\textbf{Projection from 3D to 2D.}
After the 3D completion process, we obtain $\tilde{p}_{dense} = (\tilde{x}_i, \tilde{y}_i, \tilde{z}_i)$, and then we project each point back to the original 2D plane as $(\tilde{u}_i, \tilde{v}_i)$ with the updated depth value $\tilde{d}_i=\tilde{z}_i-\varepsilon$ (we ignore the $\varepsilon$ for simplicity in the main paper) by the inverse operation of Eq.~\eqref{equ:project2to3}:
\begin{equation}
\label{equ:project3to2}
\tilde{u_i}=\frac{{}\tilde{x_i} \cdot  f}{\tilde{z_i}}+o_x, \tilde{v_i}=\frac{{}\tilde{y_i}\cdot f}{\tilde{z_i}}+o_y, \tilde{d_i}=\tilde{z_i}-\varepsilon,
\end{equation}
where $(\tilde{u}_i, \tilde{v}_i)$ are rounded to integers.
Since all the 3D points are generated through the completion model, the position $(\tilde{u}_i, \tilde{v}_i)$ projected from point cloud may be duplicated (i.e., two projected points happen to overlap in the 2D plane) or out of the original image plane (i.e., $(\tilde{u}_i, \tilde{v}_i)< 0$ or $(\tilde{u}_i, \tilde{v}_i)>(H,W)$). For duplicated points, we choose the minimum depth value among all duplicated points as the final depth. For those positions out of the original image plane, we view them as failing projection and do not take them as the pseudo-label $\hat{y}_{comp}$.
}

\subsection{Discussions of 2D/3D Pseudo-label}
{
We discuss whether $\hat{y}_{comp}$ is complementary to the original pseudo label $\hat{y}_{cons}$.
We observe that in Figure 3 of the main paper, for some regions that look similar, the values of the same pixel between $\hat{y}_{cons}$ and $\hat{y}_{comp}$ are very close but have a little scale shift $(<10^{-1})$. For the areas that appear different (e.g., bottom-left area), $\hat{y}_{comp}$ has nearer depth values than original $\hat{y}_{cons}$, in which nearer depth values are more reasonable for the object and grass in the bottom-left corner of the image. It shows that the completion model refers to the 3D structural information to produce better results.
}

\begin{figure*}[ht]
	\centering
    \hspace{-3.5mm}
	\stackunder[5pt]{{
    \includegraphics[width=0.33\textwidth]{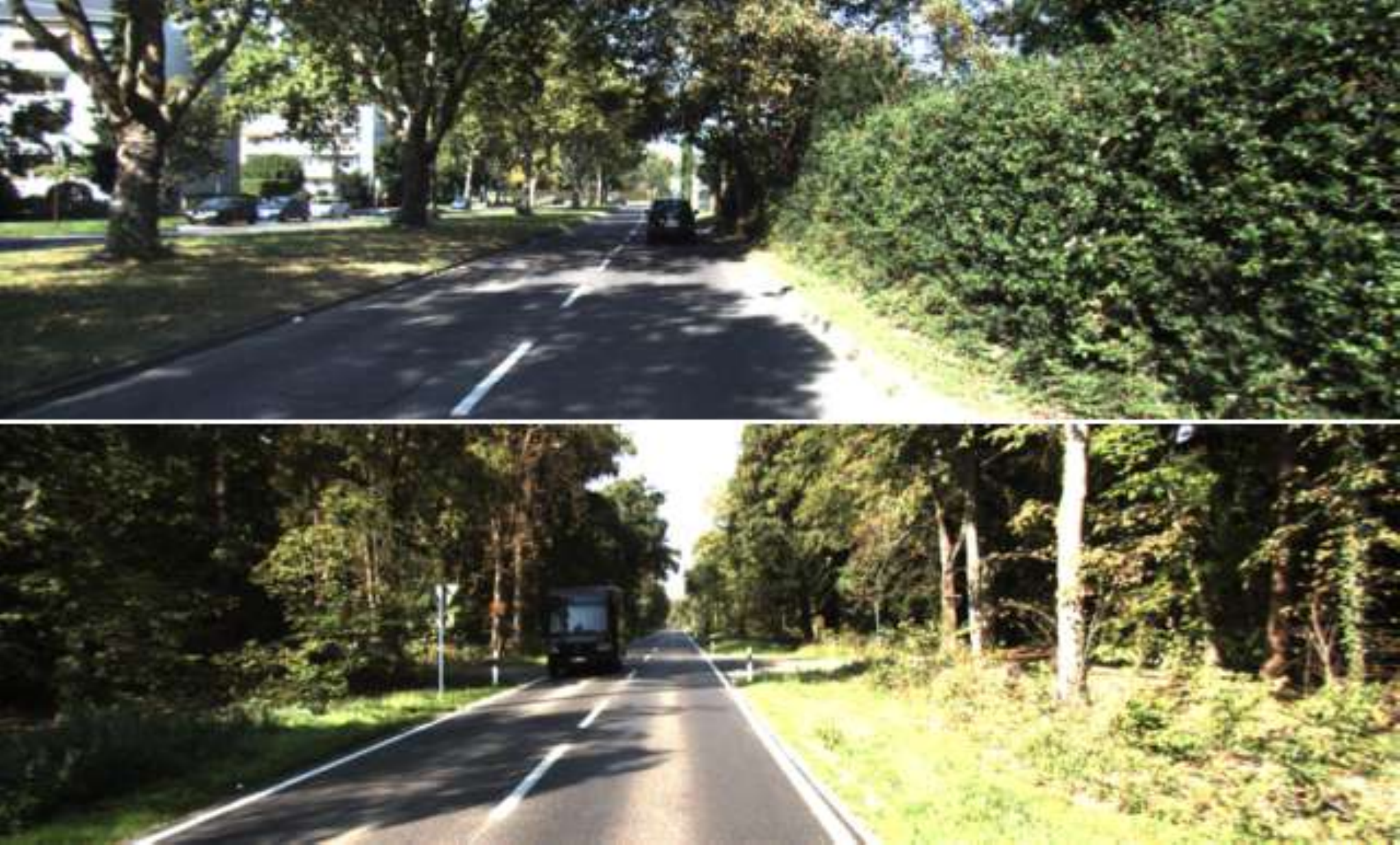}}}{Input image $x_r$}
    \hspace{-3mm}
    \stackunder[5pt]{
    \includegraphics[width=0.33\textwidth]{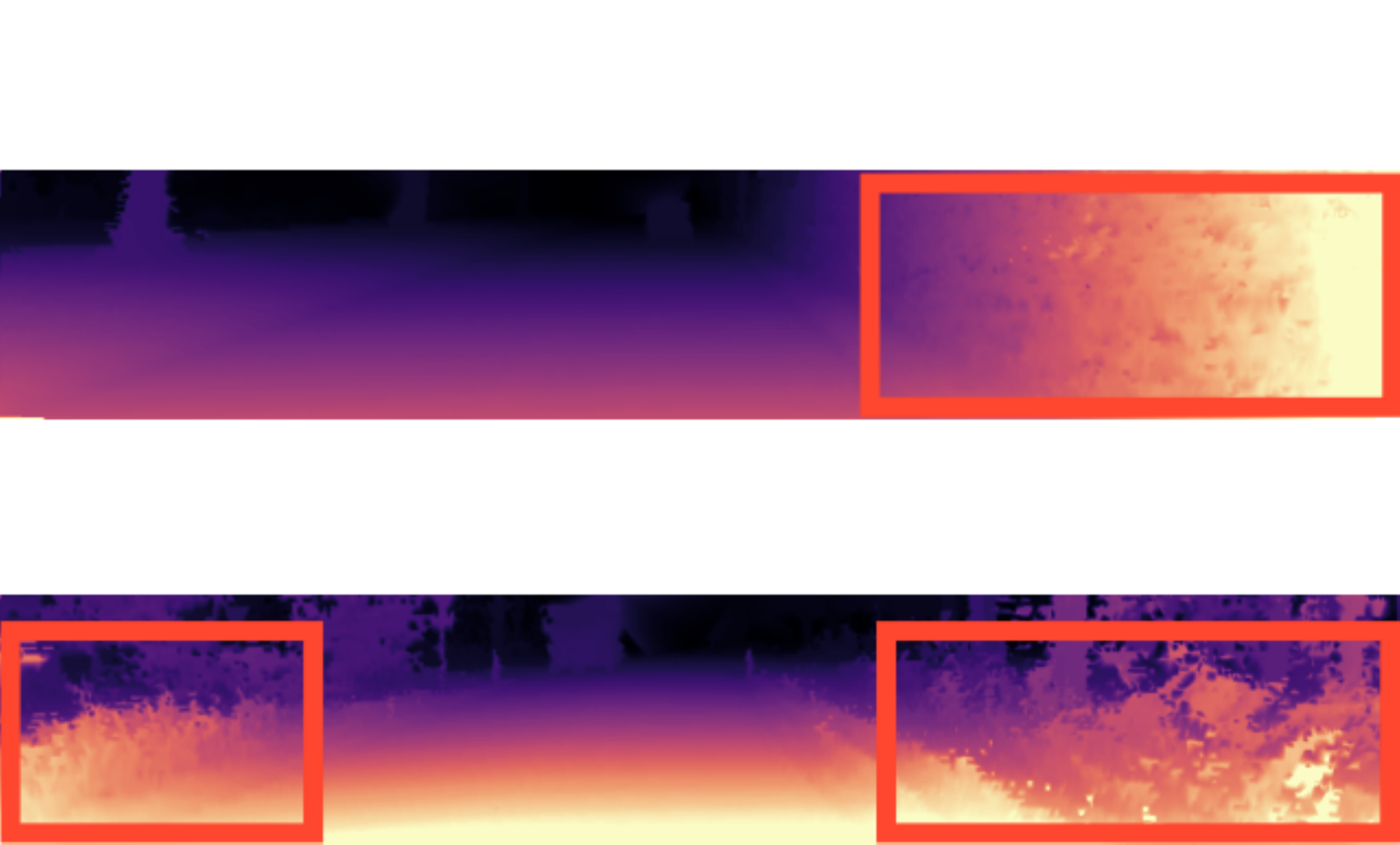}}{Ground truth ${y_r}$}
    \hspace{-3mm}
    \stackunder[5pt]{
    \includegraphics[width=0.33\textwidth]{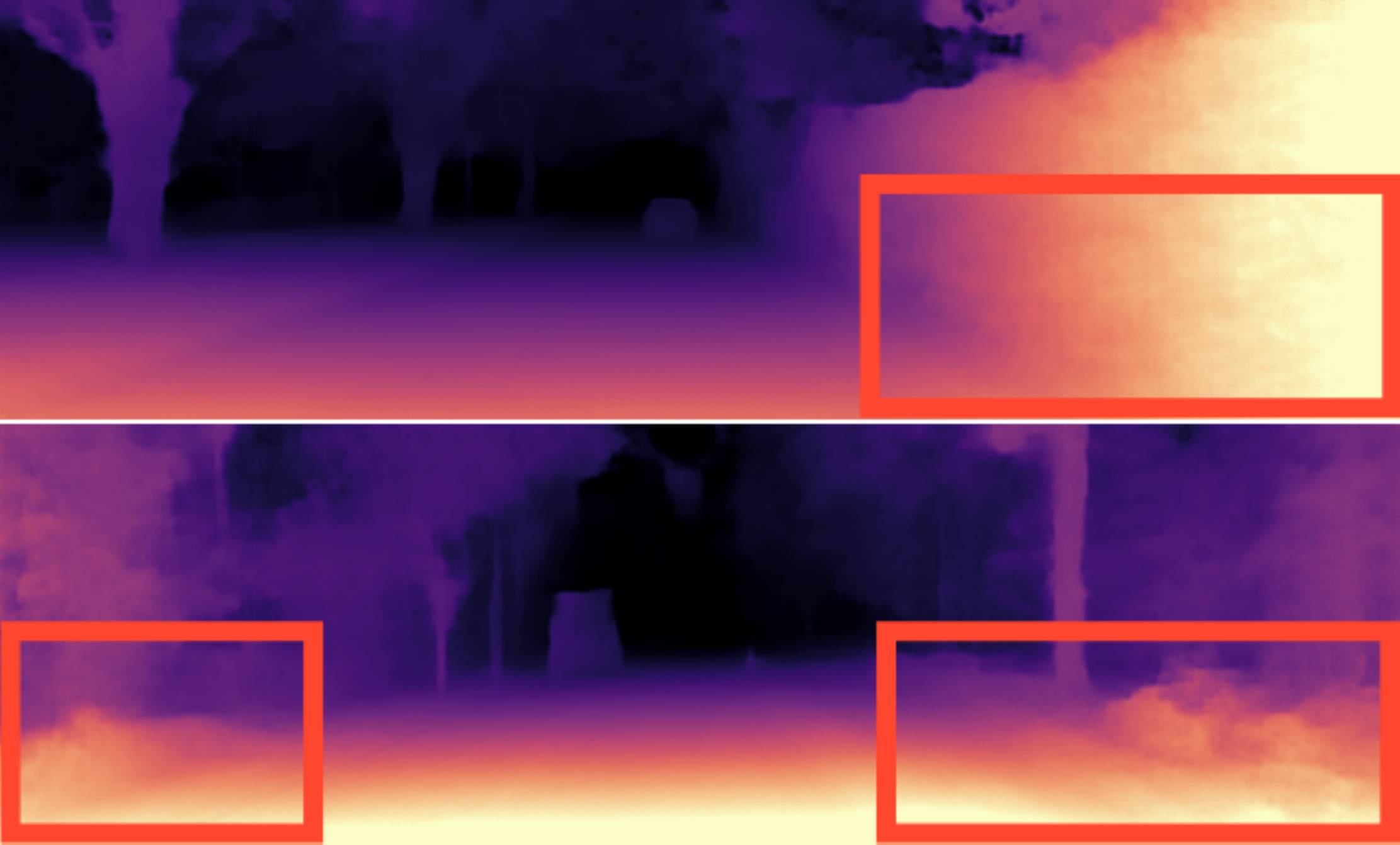}}{3D-PL + Stereo}
    \hspace{-3mm}
    \vspace{-2mm}
    \caption{3D-PL produces better results in overall structure and shape of objects, but may lose some details for the objects with complicated textures such as grass and plants.
    }
	\label{fig:limitation}
	\vspace{-2mm}
\end{figure*}

\section{Limitations}
\label{sec:limitations}
{Figure~\ref{fig:limitation} shows one example of the limitation in our 3D-PL with the stereo-pair setting. Since 3D-PL focuses on the structural information, it can perform well on the overall structure, e.g., the shape of cars and the hard objects such as road signs or traffic lights.
However, for the object that has complicated textures like grass, 3D-PL produces smoother results but loses the details of the plant.

}

\section{More qualitative results}
\label{sec:qualitative_supp}
\begin{figure*}[!t]
	\centering
    \hspace{-5mm}
	\stackunder[5pt]{{
    \includegraphics[width=0.25\textwidth]{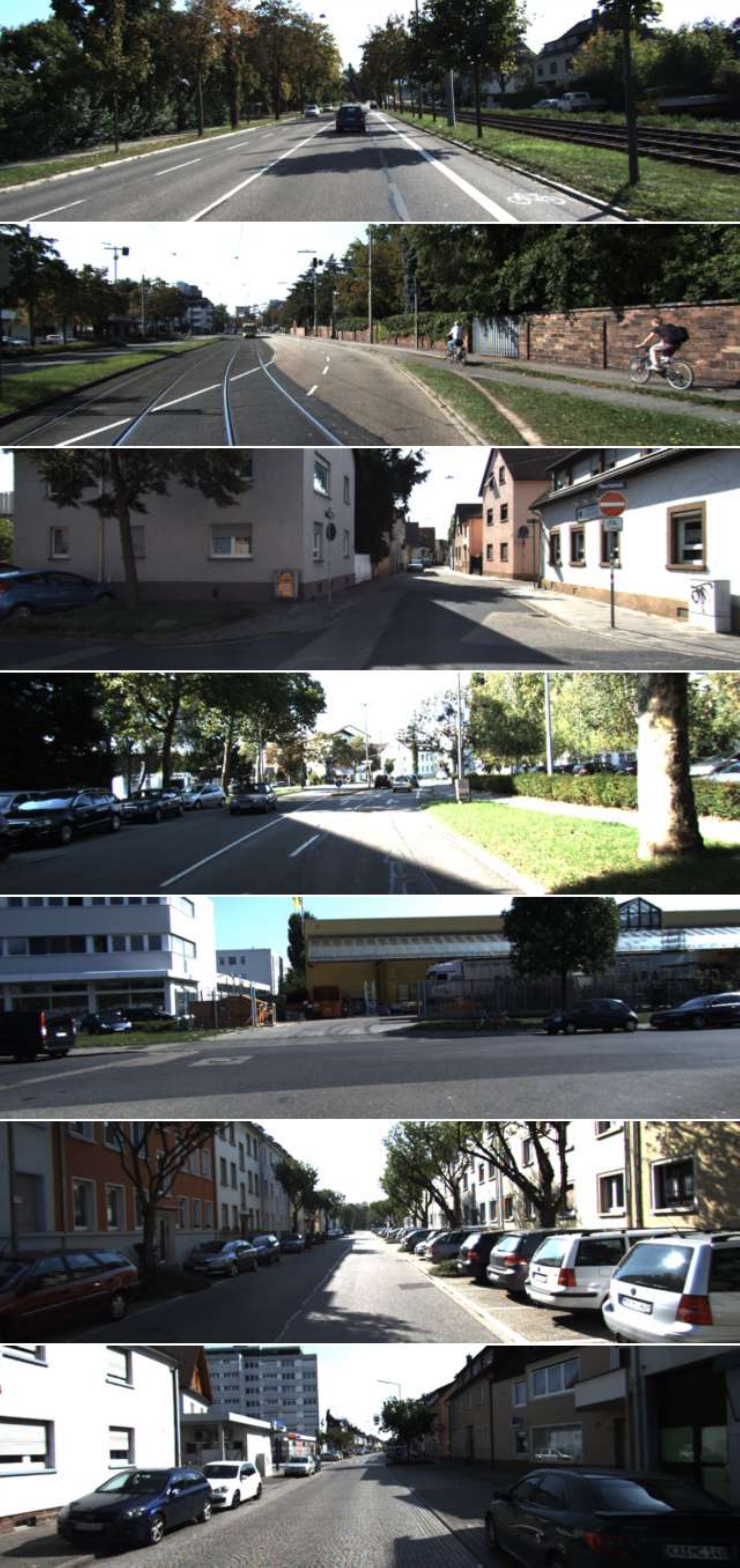}}}{Input image $x_r$}
    \hspace{-3mm}
    \stackunder[5pt]{
    \includegraphics[width=0.25\textwidth]{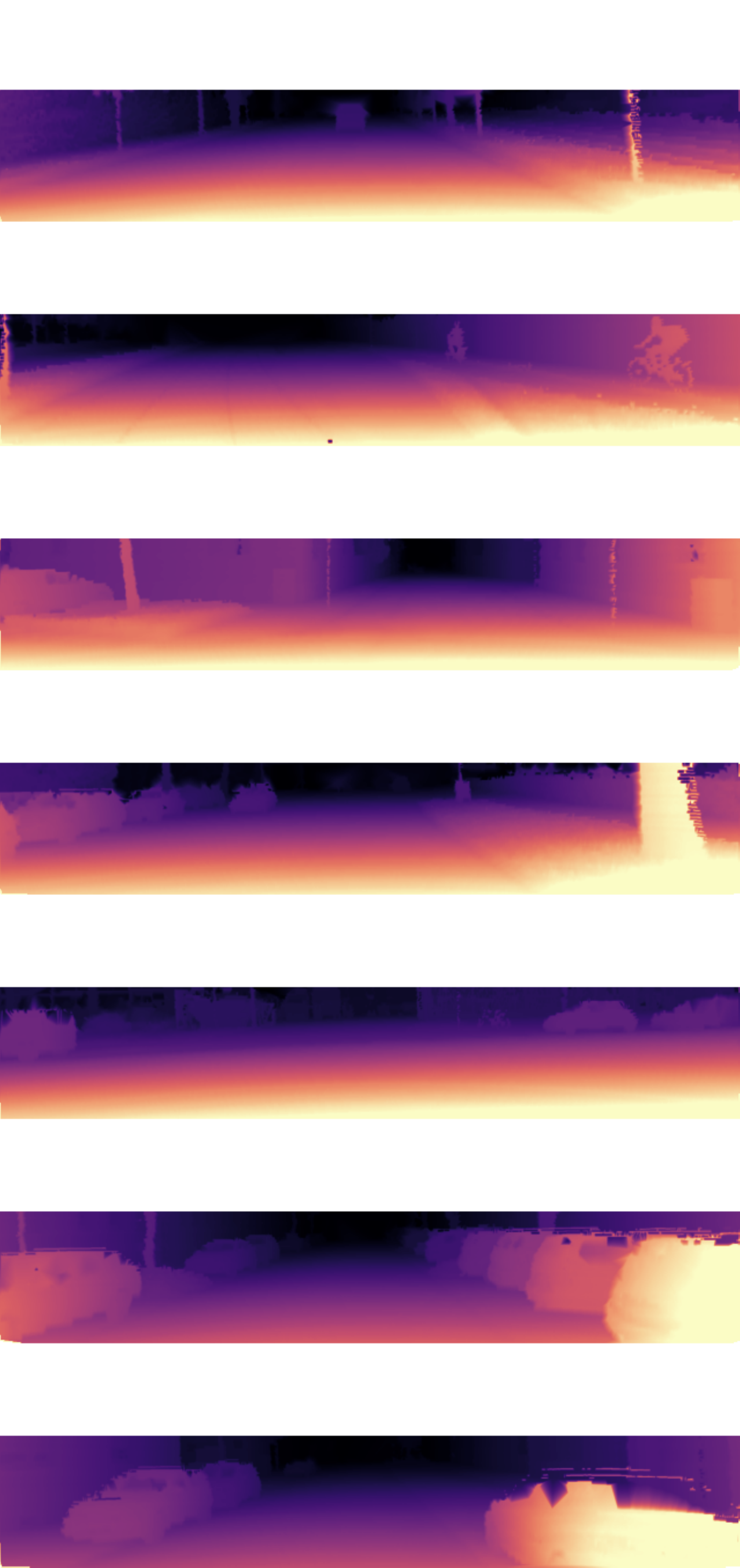}}{Ground truth ${y_r}$}
    \hspace{-3mm}
    \stackunder[5pt]{
    \includegraphics[width=0.25\textwidth]{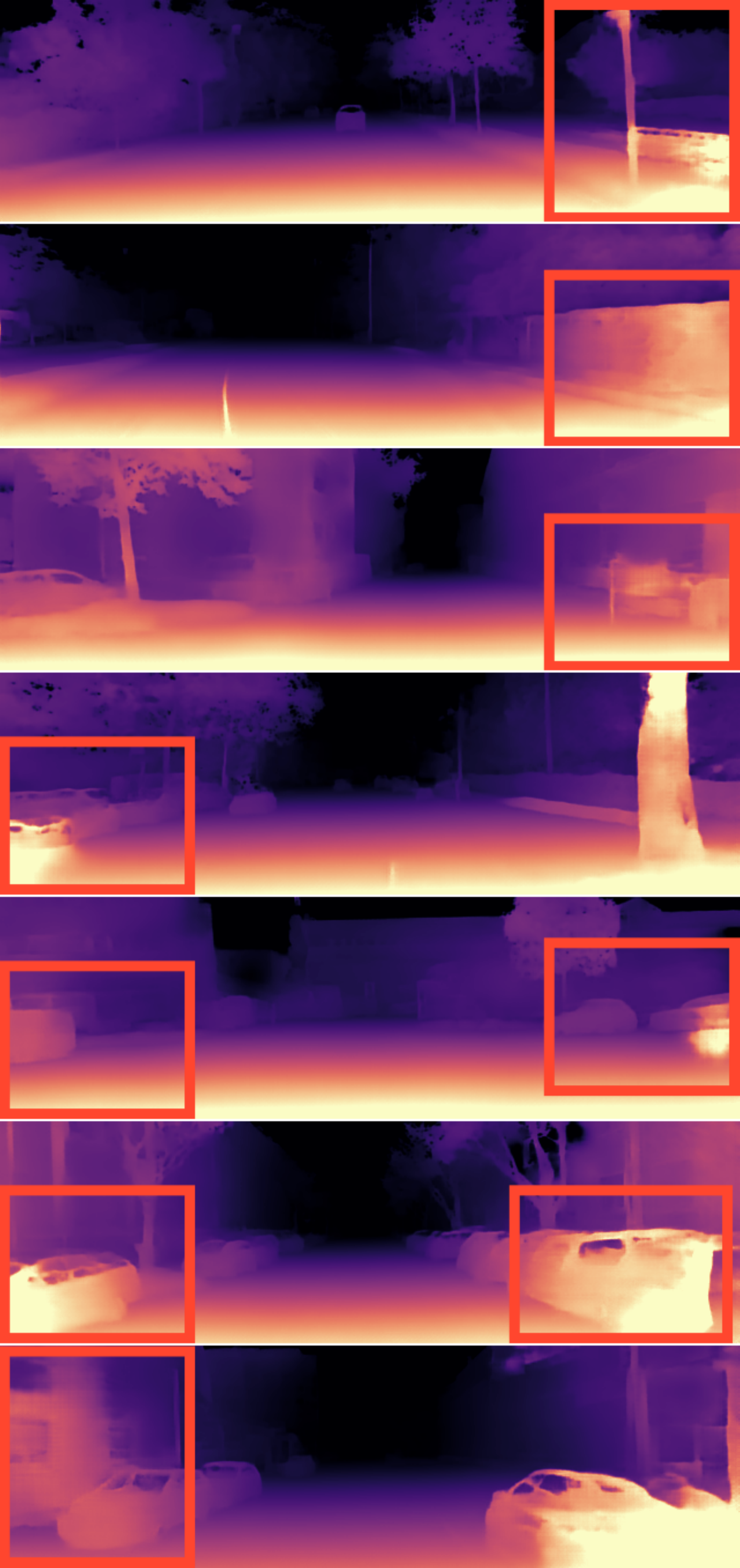}}{T$^2$Net~\cite{zheng2018t2net}
}
    \hspace{-3mm}
    \stackunder[5pt]{
    \includegraphics[width=0.25\textwidth]{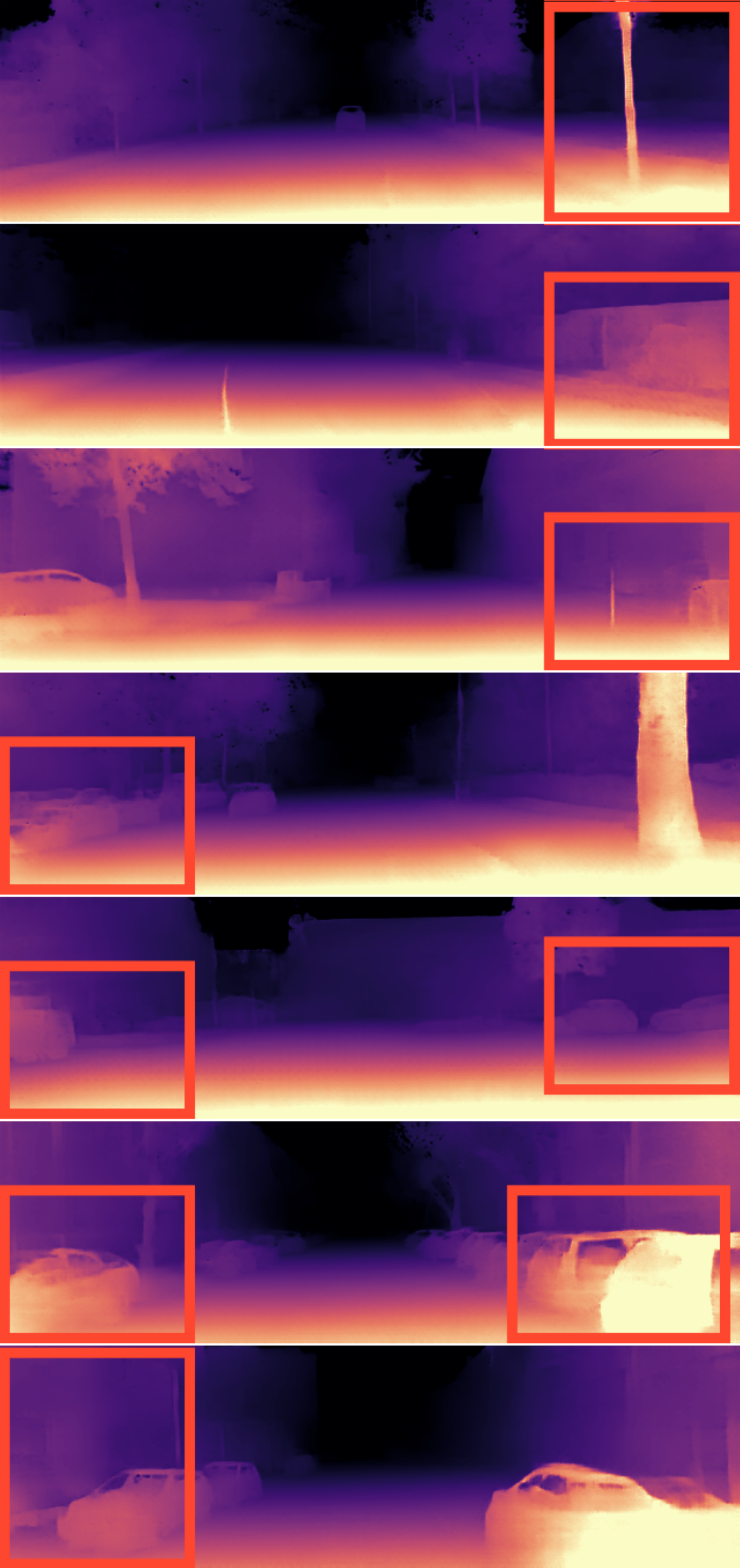}}{3D-PL}
    \hspace{-5mm} 
    \vspace{-2mm}
    \caption{More qualitative results on KITTI~\cite{geiger2012we} in the single-image setting.
    }
	\label{fig:qualitative_no_stereo_supp}
	\vspace{-3mm}
\end{figure*}

\begin{figure*}[!t]
	\centering
    \hspace{-5mm}
	\stackunder[5pt]{{
    \includegraphics[width=0.2\textwidth]{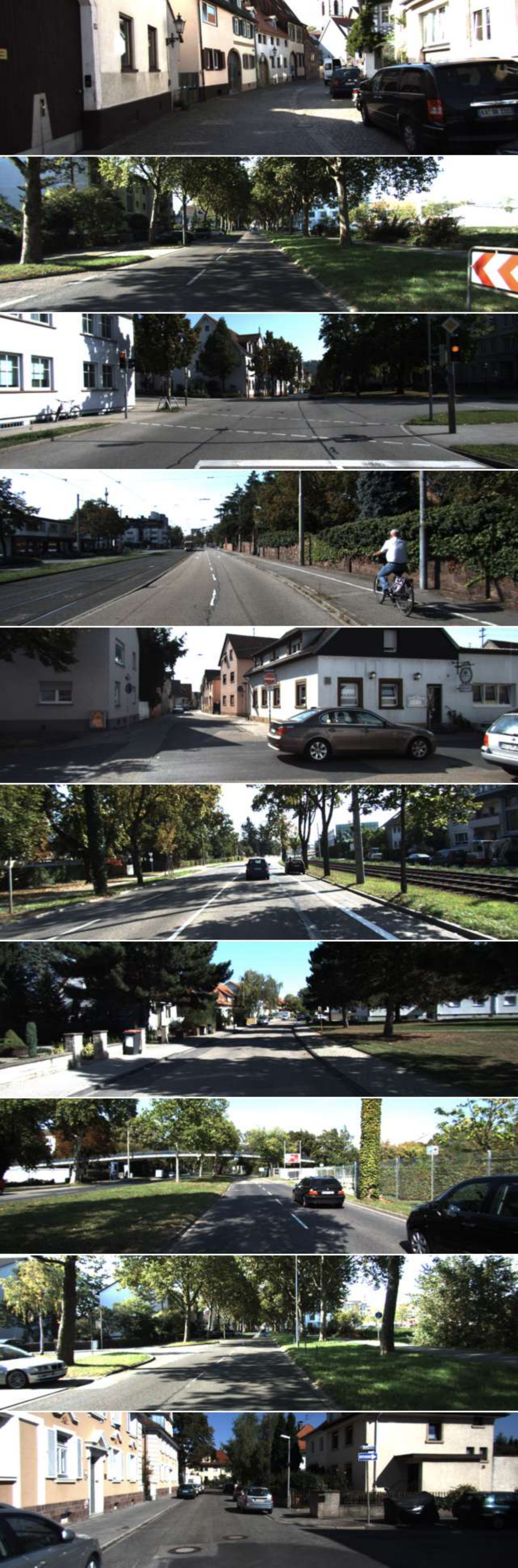}}}{Input image $x_r$}
    \hspace{-3mm}
    \stackunder[5pt]{
    \includegraphics[width=0.2\textwidth]{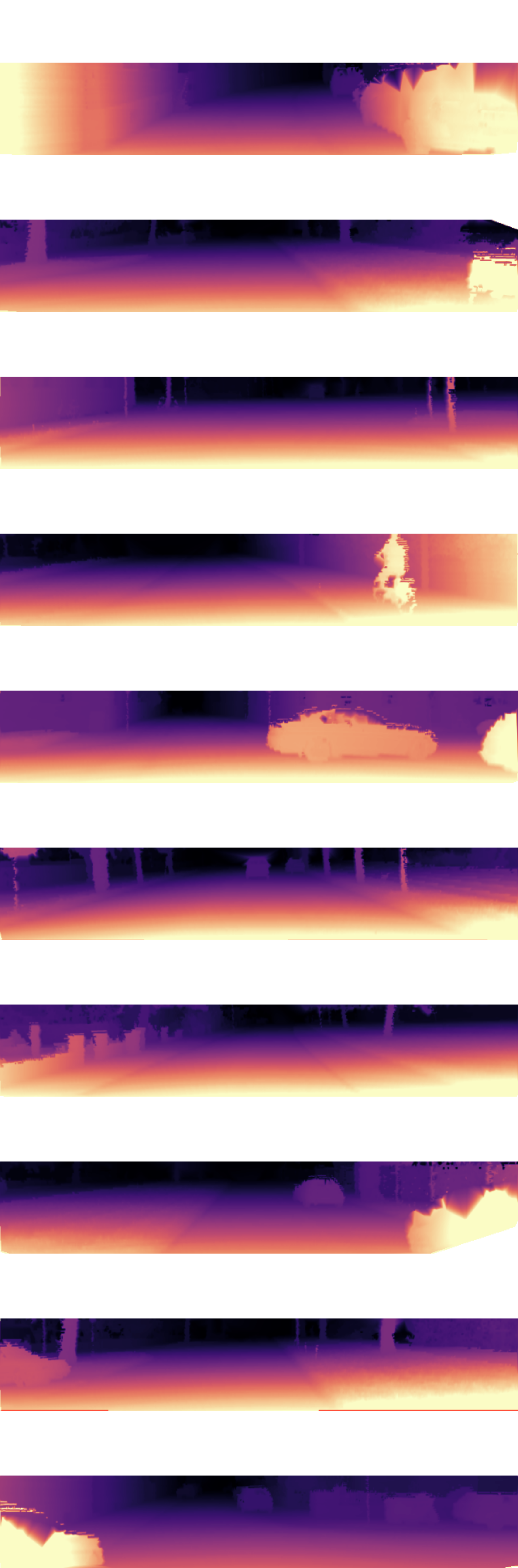}}{Ground truth ${y_r}$}
    \hspace{-3mm}
    \stackunder[5pt]{
    \includegraphics[width=0.2\textwidth]{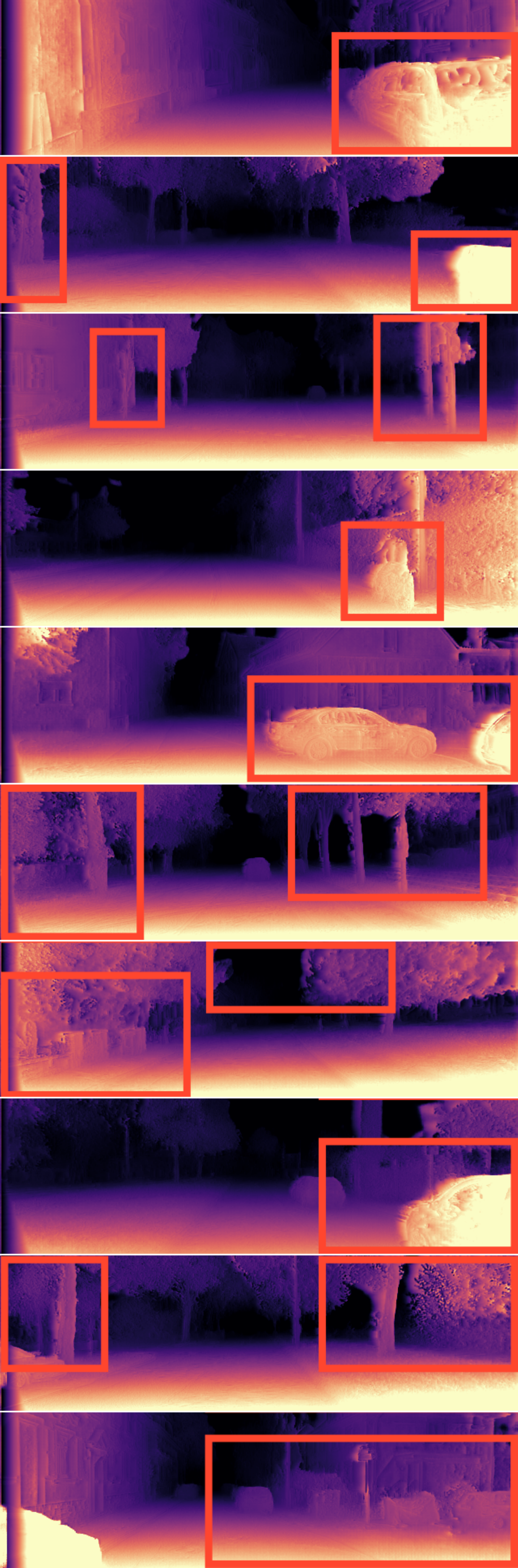}}{DESC~\cite{lopez2020desc}
}
    \hspace{-3mm}
    \stackunder[5pt]{
    \includegraphics[width=0.2\textwidth]{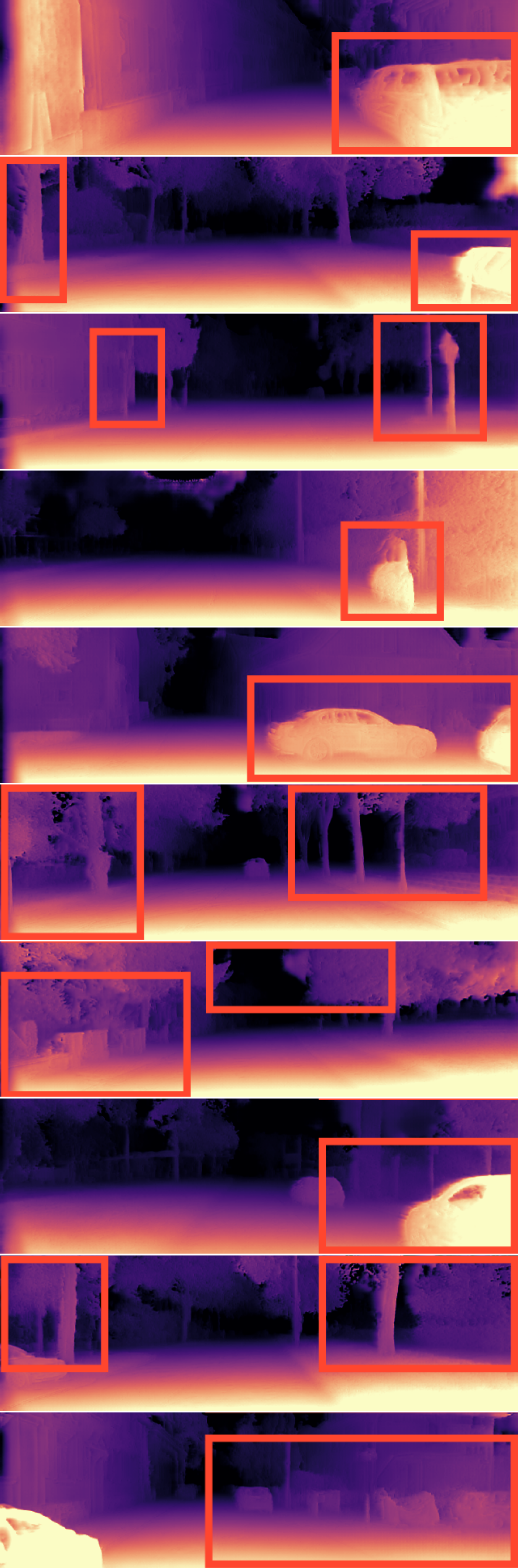}}{SharinGAN~\cite{pnvr2020sharingan}}
    \hspace{-3mm}
    \stackunder[5pt]{
    \includegraphics[width=0.2\textwidth]{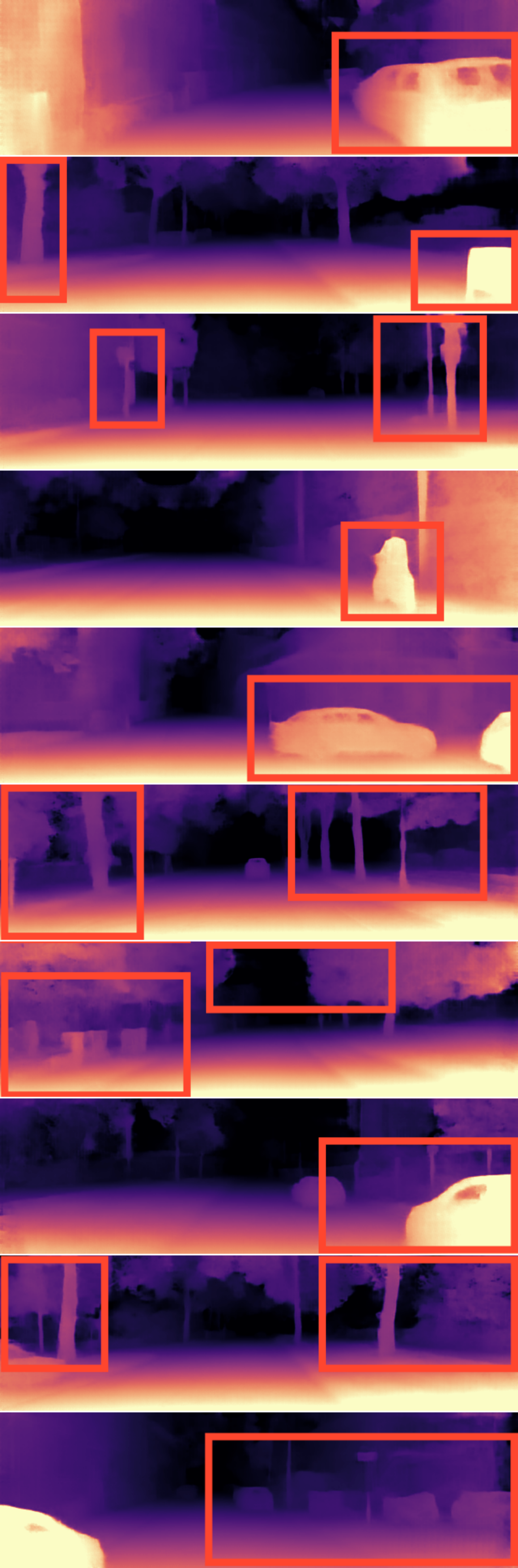}}{3D-PL+Stereo}
    \hspace{-5mm}
    \vspace{-2mm}
    \caption{More qualitative results on KITTI~\cite{geiger2012we} with having stereo pairs during training.
    }
	\label{fig:qualitative_stereo_supp}
	\vspace{-3mm}
\end{figure*}

\begin{figure*}[!t]
	\centering
    \hspace{-5mm}
	\stackunder[5pt]{{
    \includegraphics[width=0.25\textwidth]{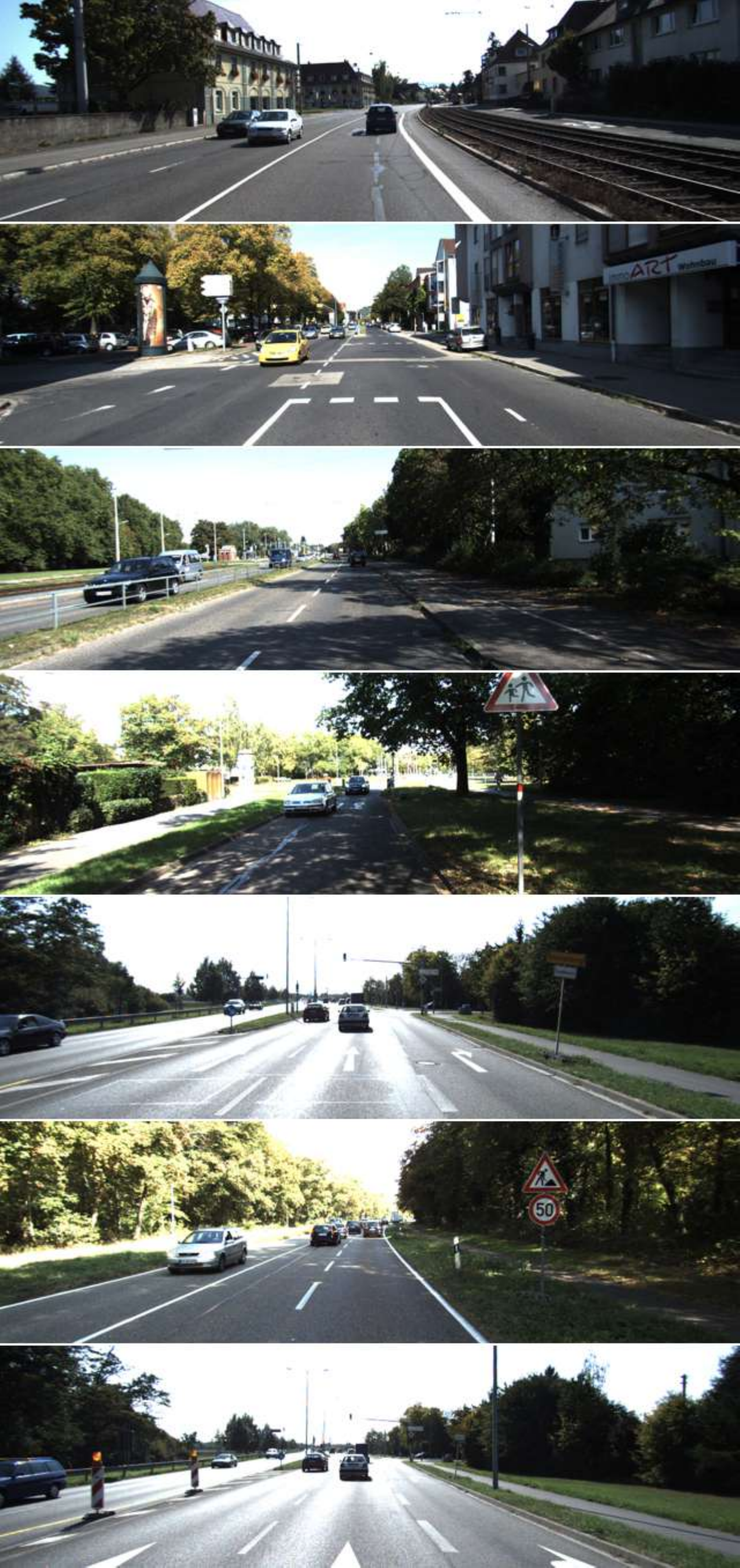}}}{Input image $x_r$}
    \hspace{-3mm}
    \stackunder[5pt]{
    \includegraphics[width=0.25\textwidth]{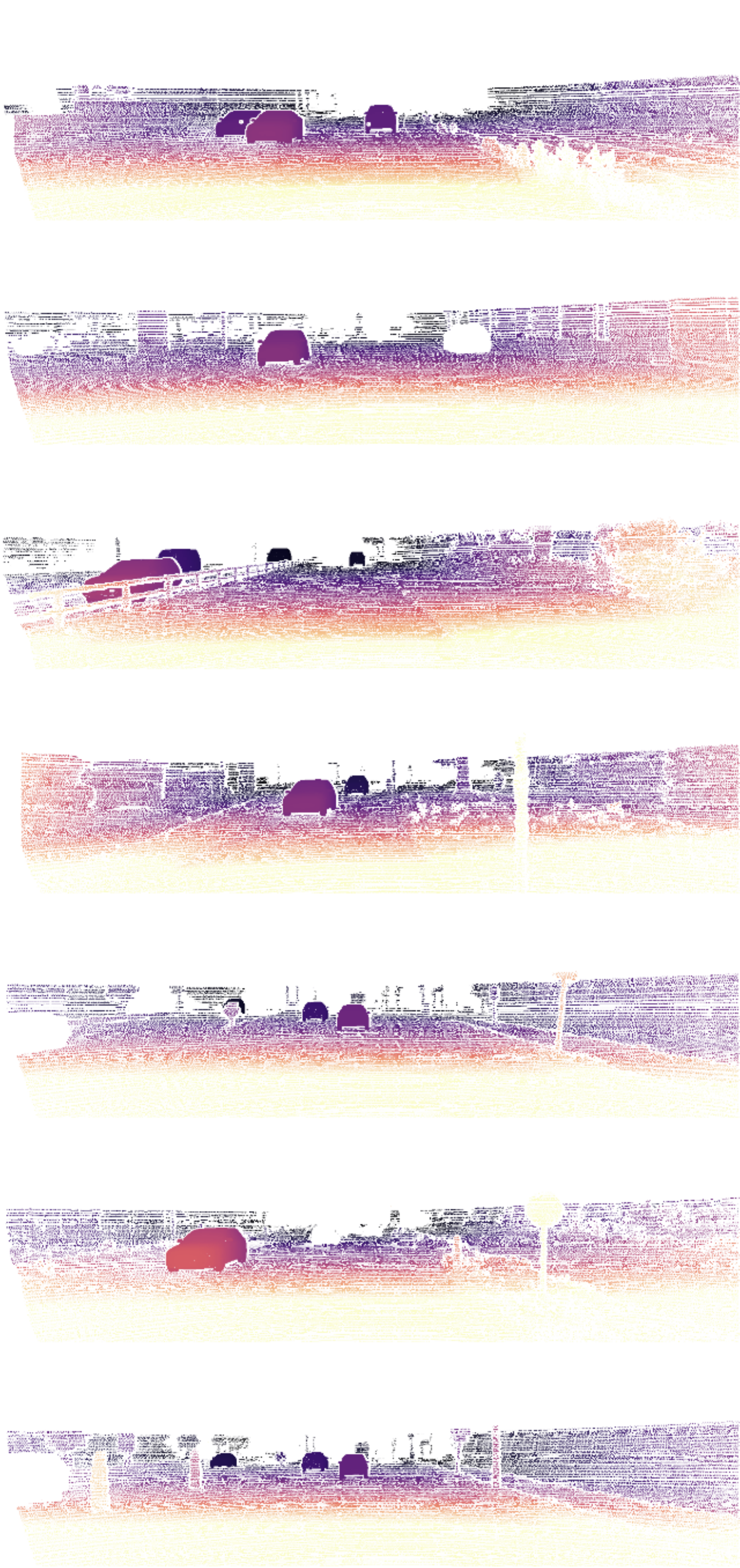}}{Ground truth ${y_r}$}
    \hspace{-3mm}
    \stackunder[5pt]{
    \includegraphics[width=0.25\textwidth]{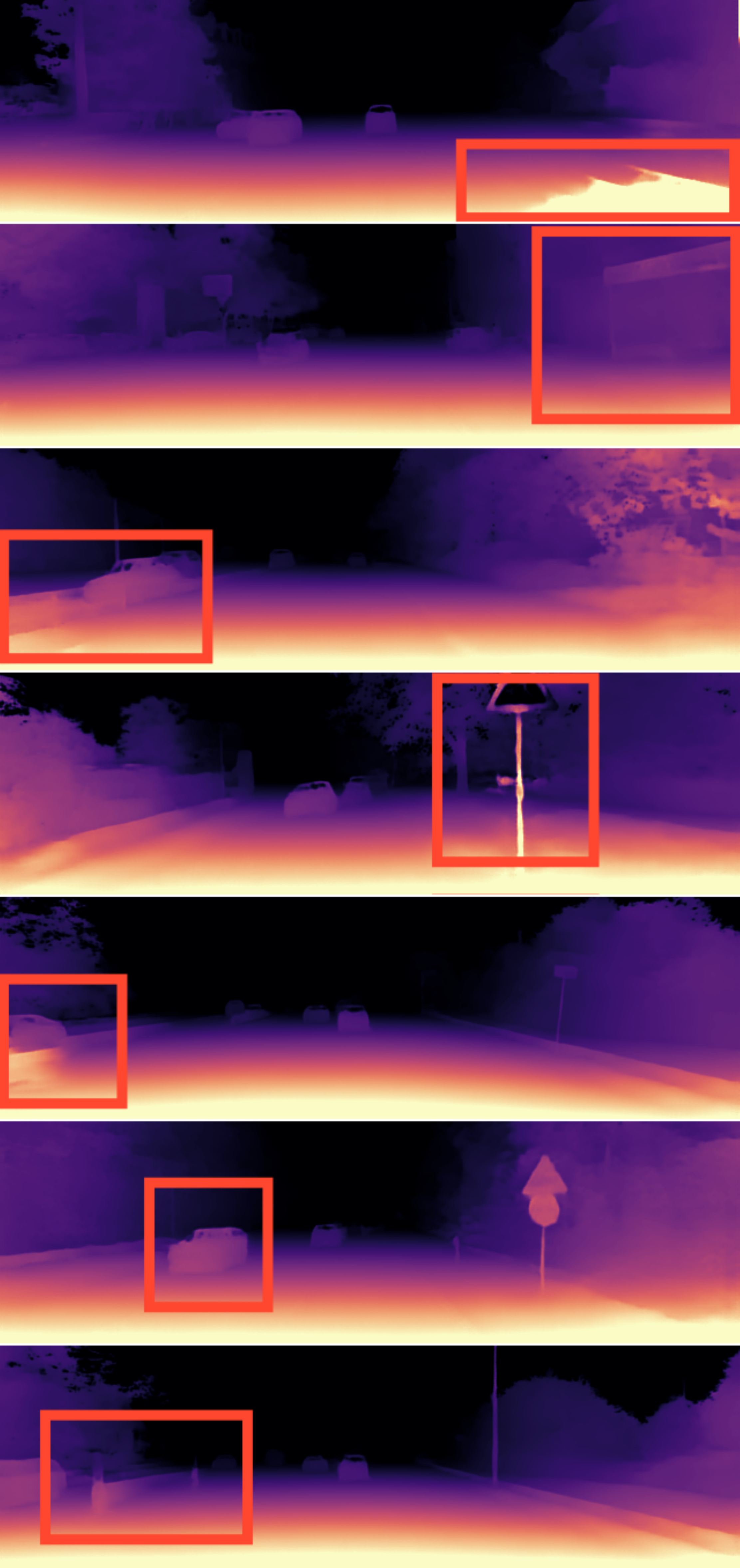}}{T$^2$Net~\cite{zheng2018t2net}
}
    \hspace{-3mm}
    \stackunder[5pt]{
    \includegraphics[width=0.25\textwidth]{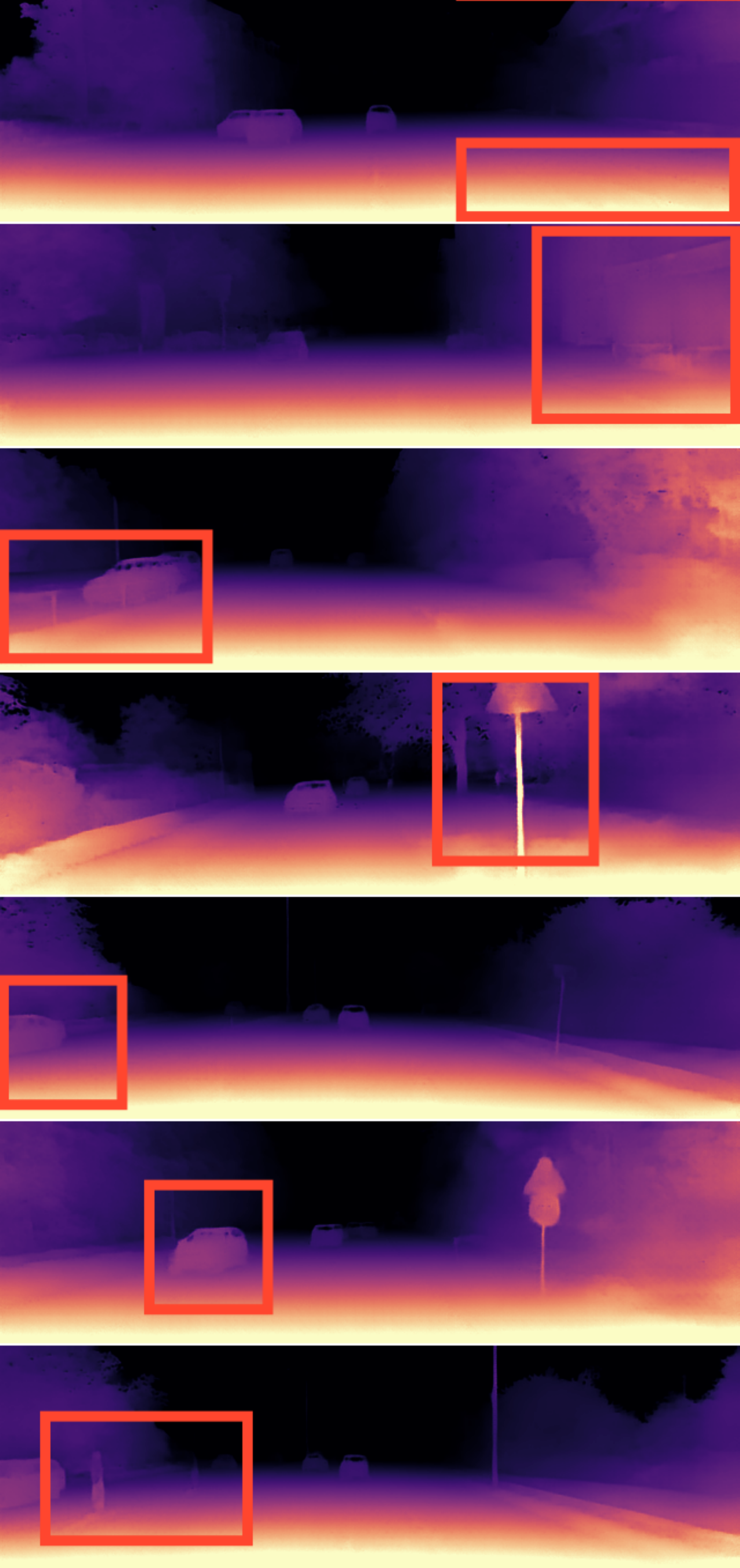}}{3D-PL}
    \hspace{-5mm} 
    \vspace{-2mm}
    \caption{More qualitative results on KITTI stereo 2015~\cite{menze2015object} in the single-image setting.
    }
	\label{fig:qualitative_2015_no_stereo}
	\vspace{-3mm}
\end{figure*}

\begin{figure*}[!t]
	\centering
    \hspace{-5mm}
	\stackunder[5pt]{{
    \includegraphics[width=0.2\textwidth]{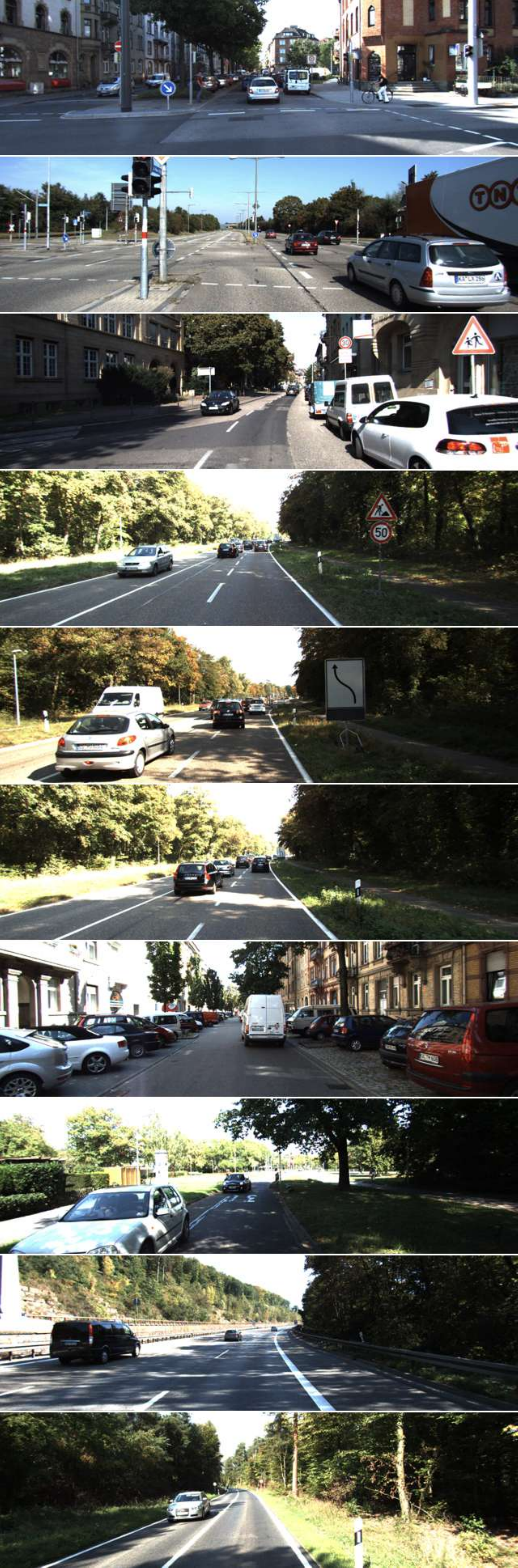}}}{Input image $x_r$}
    \hspace{-3mm}
    \stackunder[5pt]{
    \includegraphics[width=0.2\textwidth]{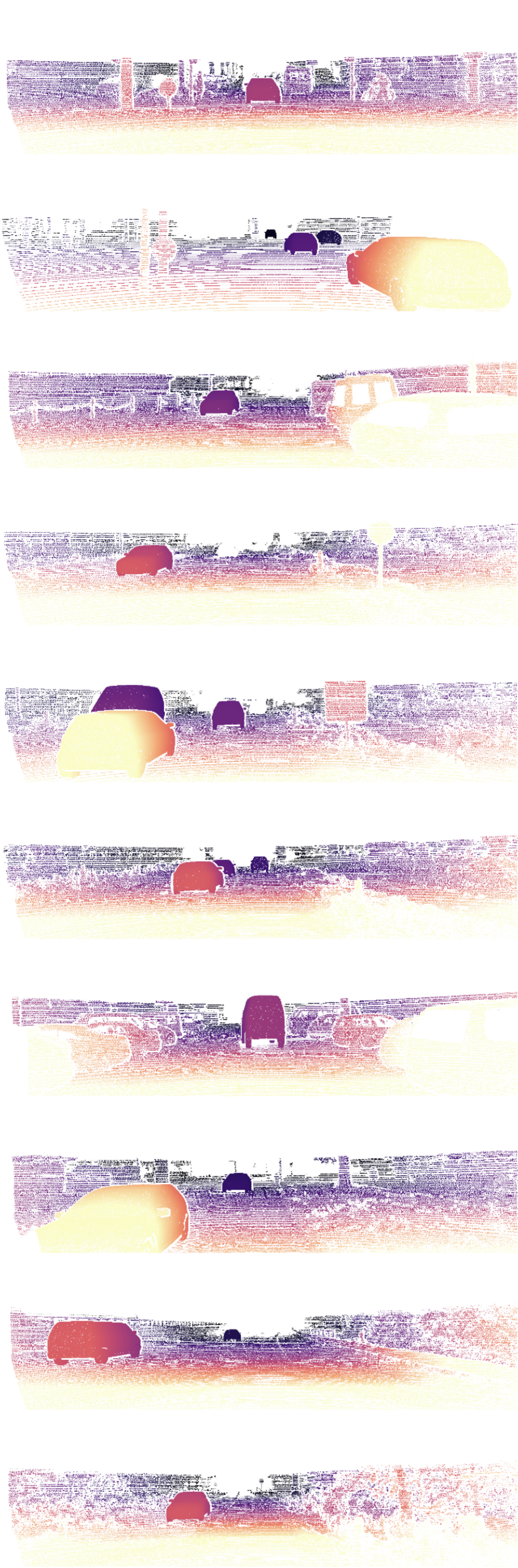}}{Ground truth ${y_r}$}
    \hspace{-3mm}
    \stackunder[5pt]{
    \includegraphics[width=0.2\textwidth]{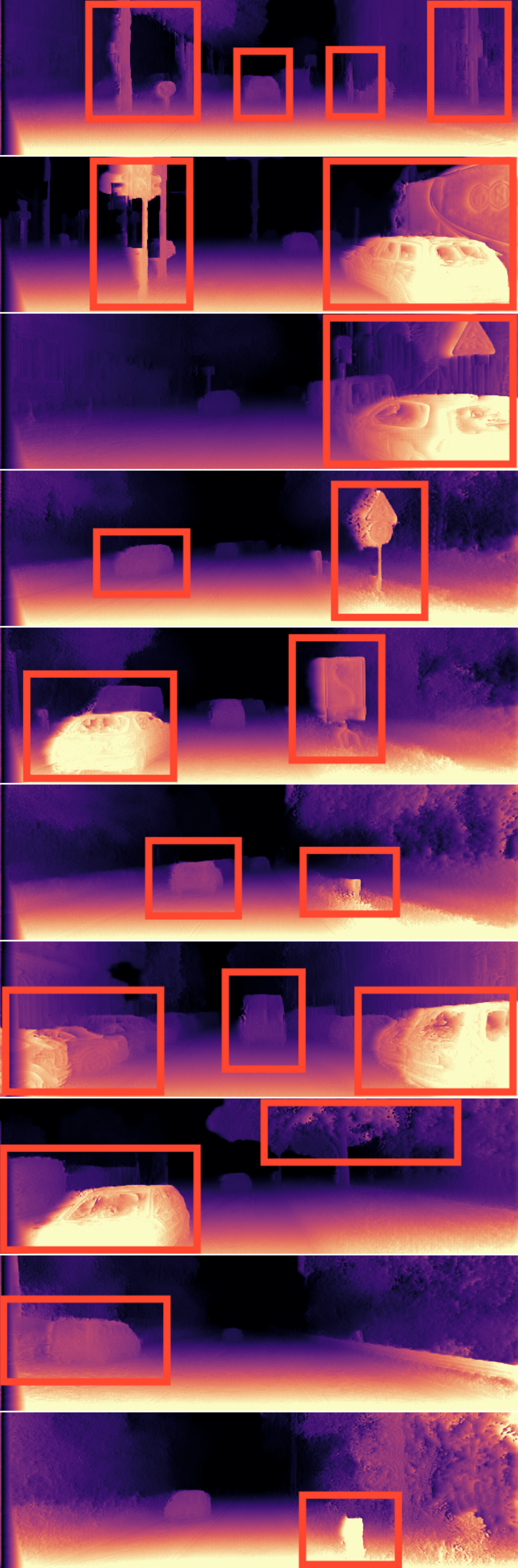}}{DESC~\cite{lopez2020desc} 
}
    \hspace{-3mm}
    \stackunder[5pt]{
    \includegraphics[width=0.2\textwidth]{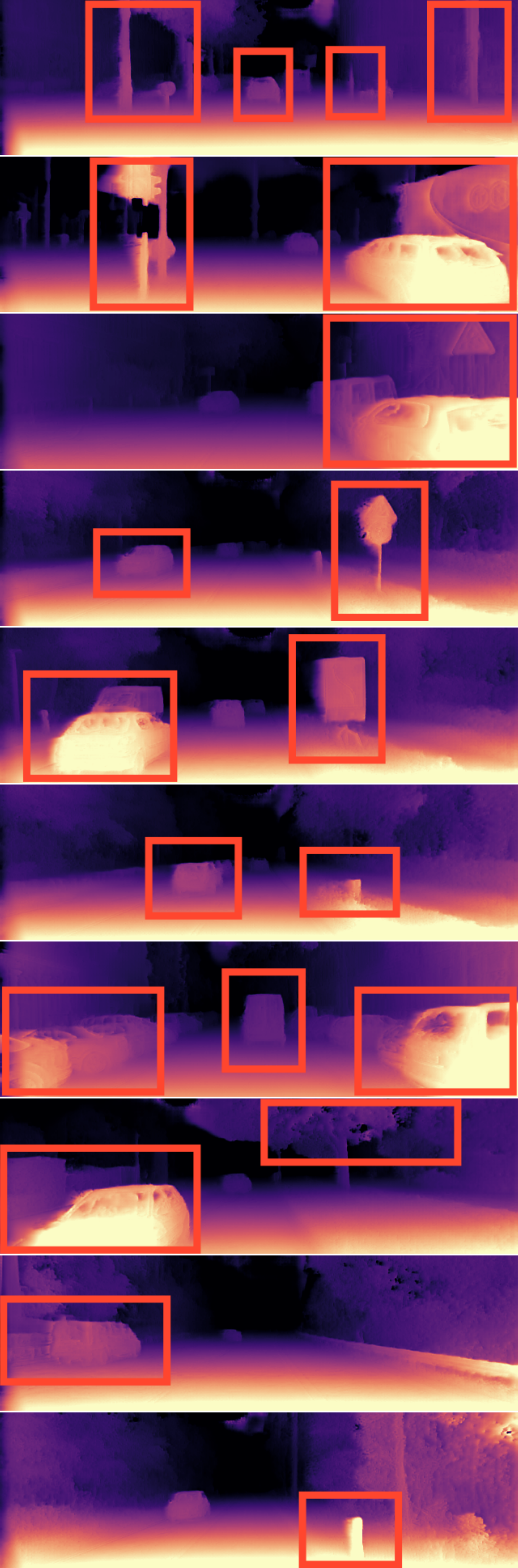}}{SharinGAN~\cite{pnvr2020sharingan}}
    \hspace{-3mm}
    \stackunder[5pt]{
    \includegraphics[width=0.2\textwidth]{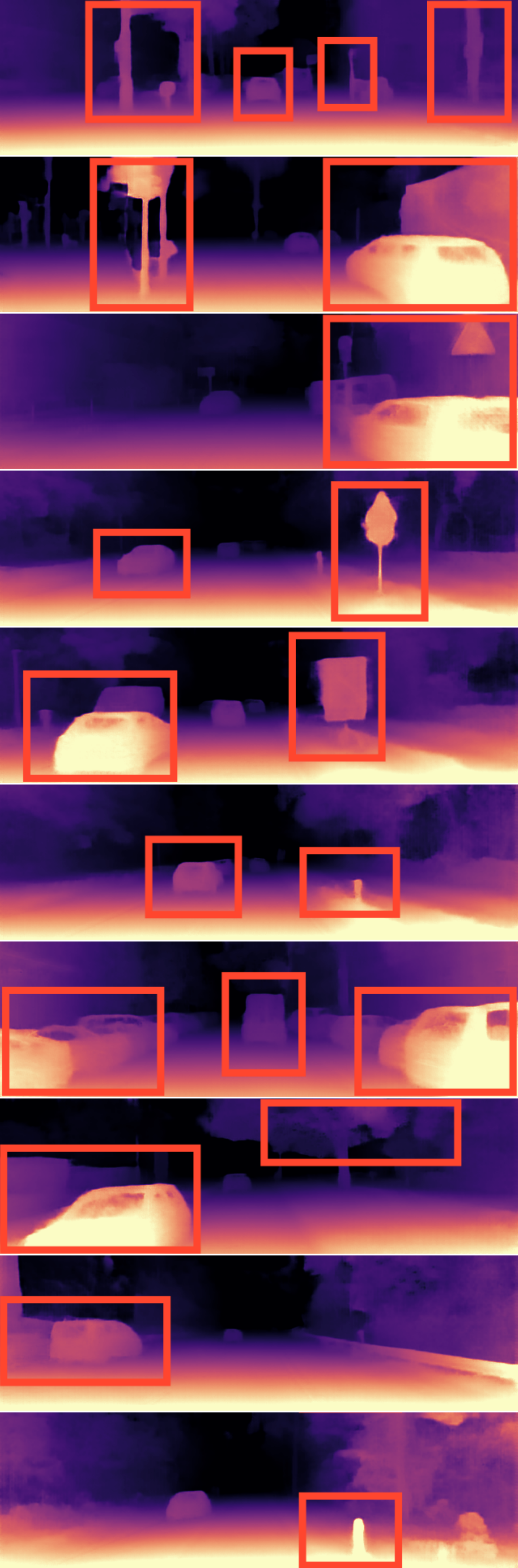}}{3D-PL+Stereo}
    \hspace{-5mm}
    \vspace{-2mm}
    \caption{More qualitative results on KITTI stereo 2015~\cite{menze2015object} with having stereo pairs during training.
    }
	\label{fig:qualitative_2015}
	\vspace{-3mm}
\end{figure*}

\begin{figure*}[!t]
	\centering
    \hspace{-5mm}
	\stackunder[5pt]{{
    \includegraphics[width=0.25\textwidth]{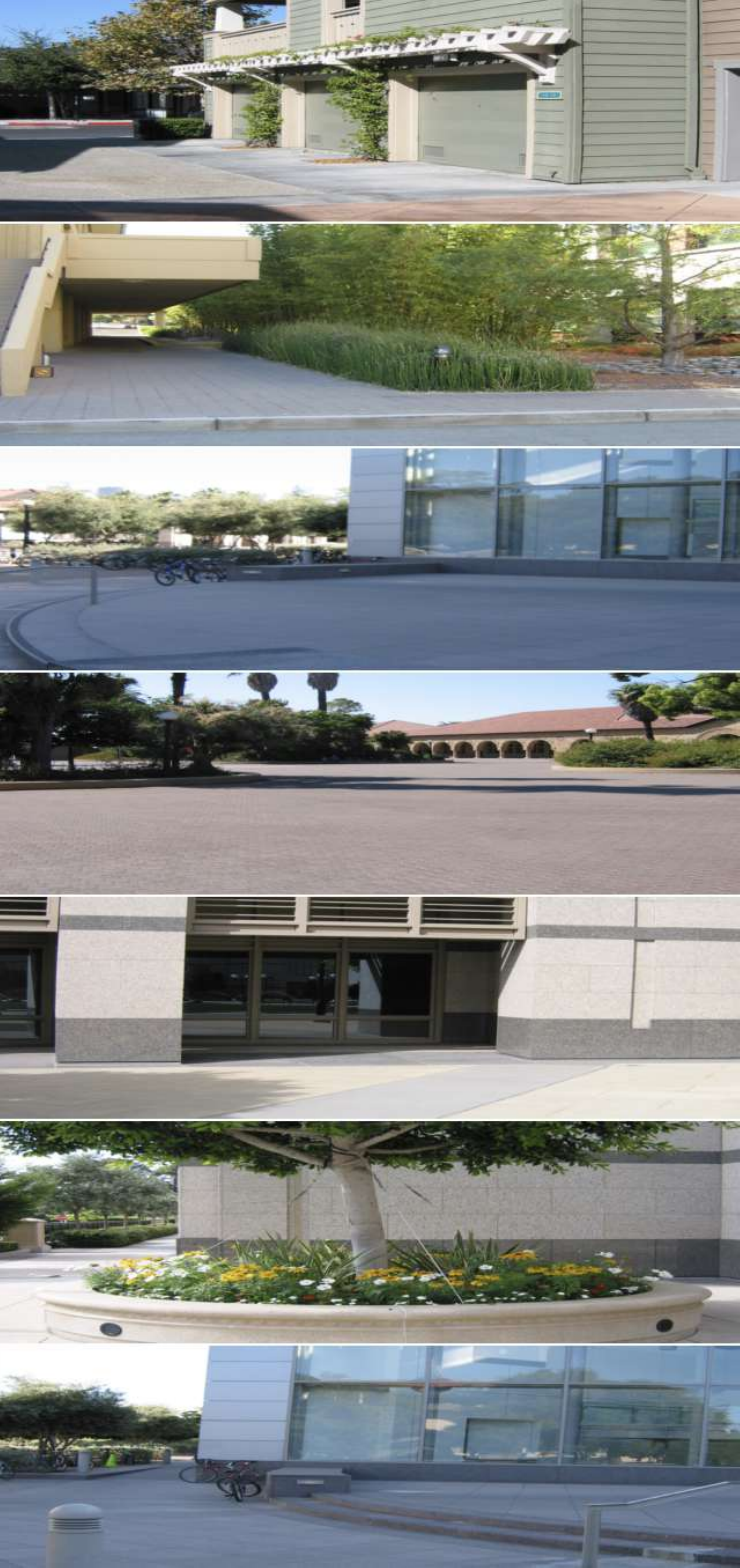}}}{Input image $x_r$}
    \hspace{-3mm}
    \stackunder[5pt]{
    \includegraphics[width=0.25\textwidth]{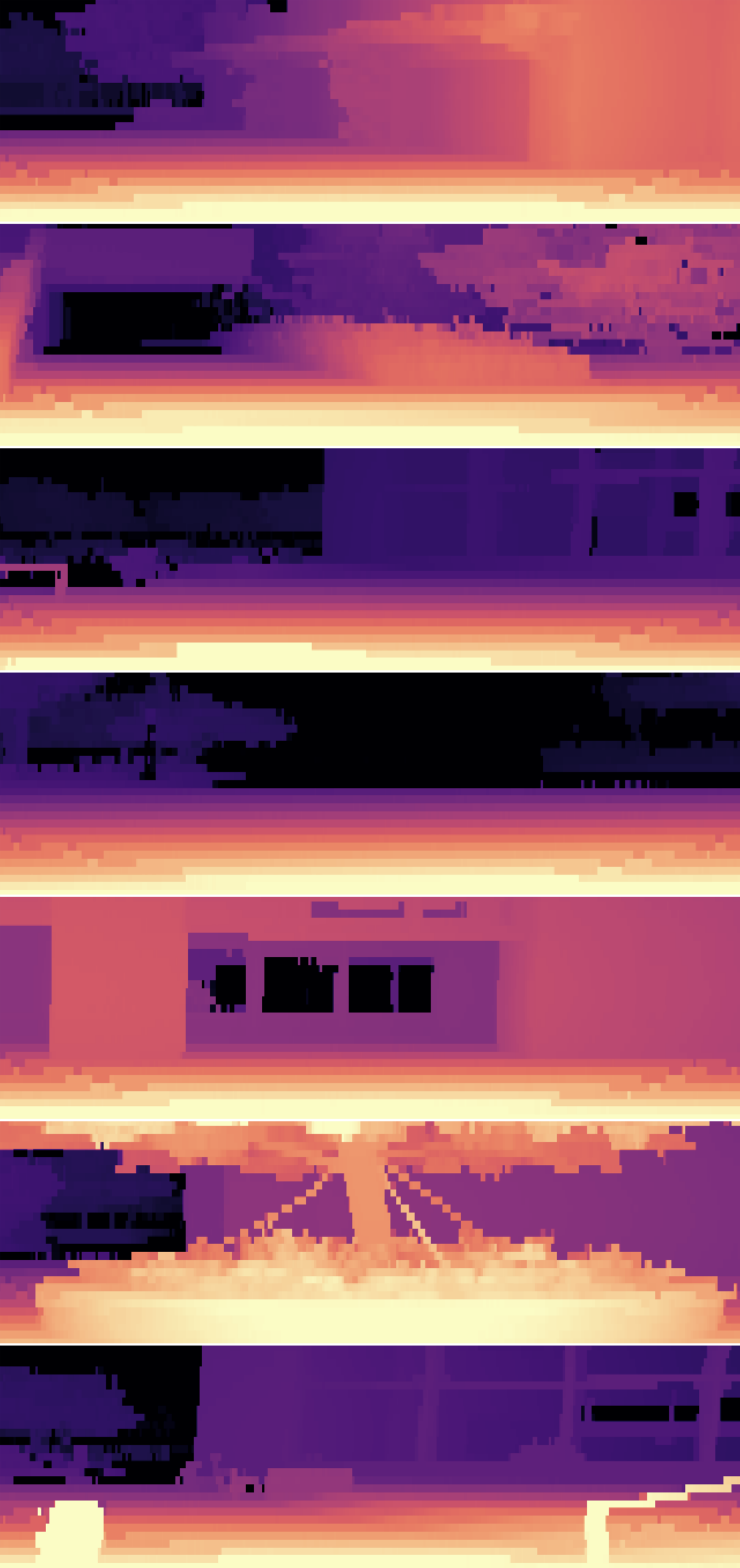}}{Ground truth ${y_r}$}
    \hspace{-3mm}
    \stackunder[5pt]{
    \includegraphics[width=0.25\textwidth]{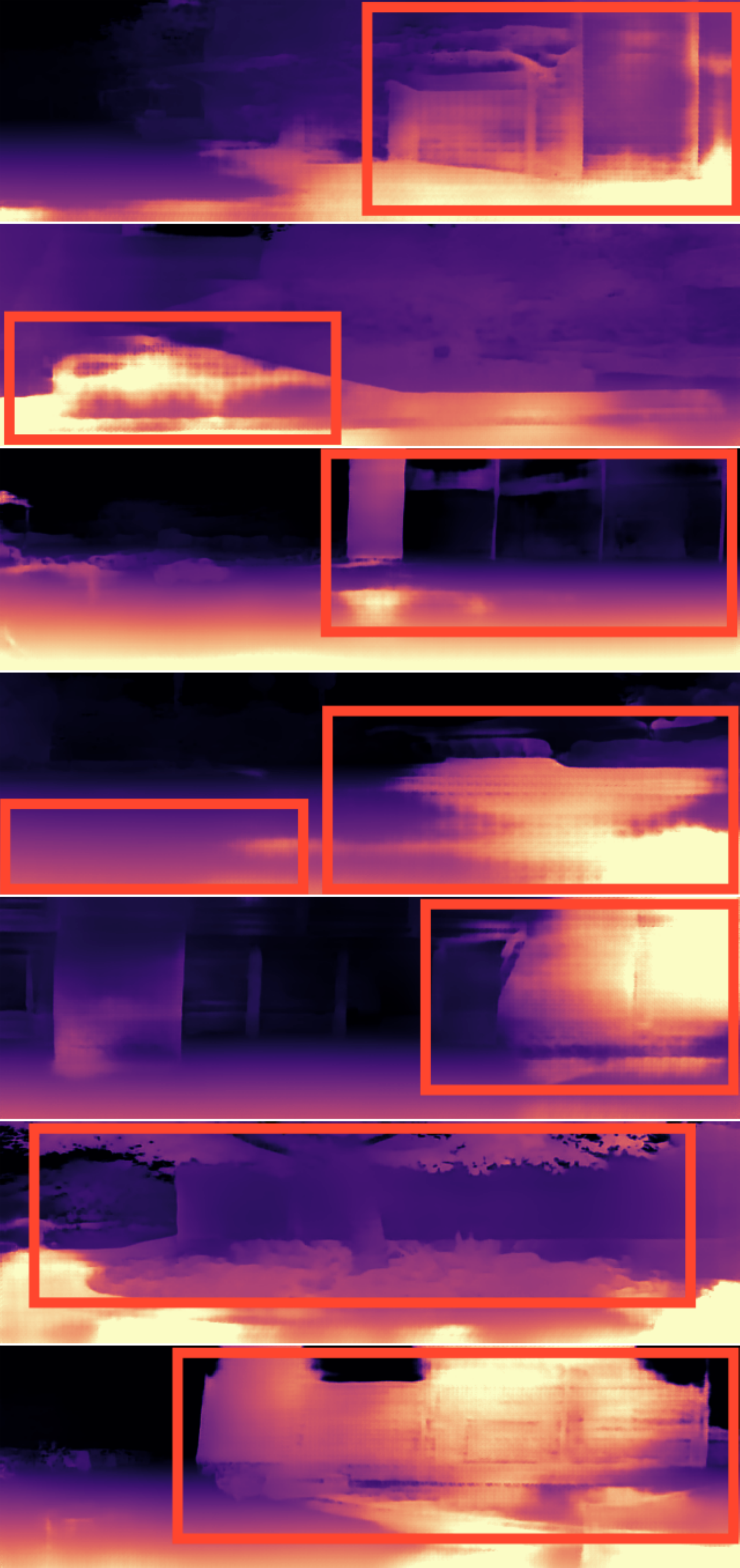}}{T$^2$Net~\cite{zheng2018t2net}
}
    \hspace{-3mm}
    \stackunder[5pt]{
    \includegraphics[width=0.25\textwidth]{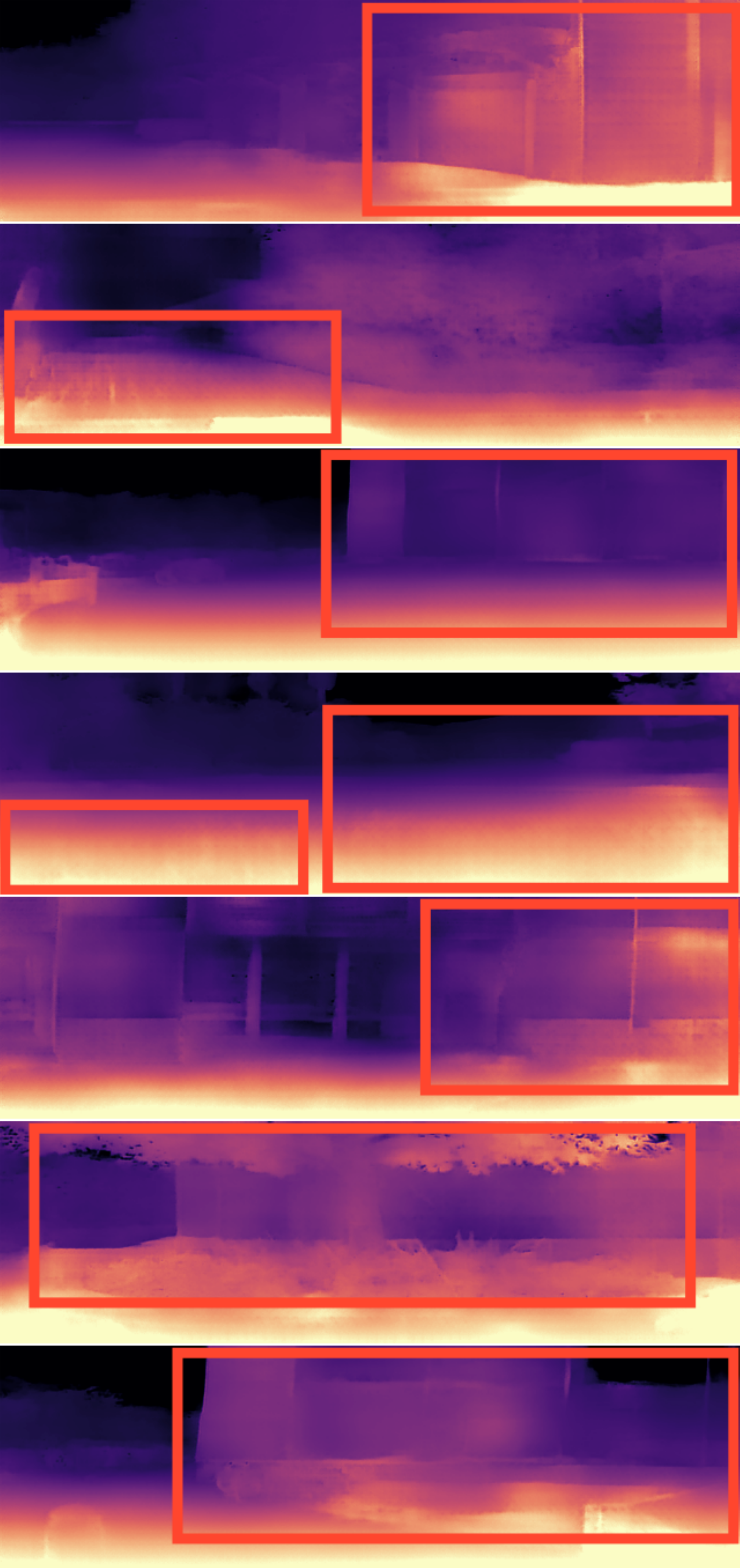}}{3D-PL}
    \hspace{-5mm} 
    \vspace{-2mm}
    \caption{More qualitative results on Make3D~\cite{saxena2008make3d} in the single-image setting.
    }
	\label{fig:qualitative_make3d}
	\vspace{-3mm}
\end{figure*}

We provide more qualitative results for different settings. Figure~\ref{fig:qualitative_no_stereo_supp} and Figure~\ref{fig:qualitative_stereo_supp} are results for KITTI~\cite{geiger2012we} in the single-image and stereo-pair settings, respectively. Figure~\ref{fig:qualitative_2015_no_stereo} and Figure~\ref{fig:qualitative_2015} are results for KITTI stereo 2015~\cite{menze2015object} in single-image and stereo-pair settings, respectively. Figure~\ref{fig:qualitative_make3d} presents results for make3D~\cite{saxena2008make3d} in the single-image setting.

\end{document}